\documentclass[runningheads]{llncs}

\usepackage{eccv}

\usepackage{eccvabbrv}

\usepackage{graphicx}
\usepackage{booktabs}
\usepackage[accsupp]{axessibility}  

\usepackage{hyperref}

\usepackage{orcidlink}

\usepackage{soul}
\usepackage[many]{tcolorbox}
\usepackage{multicol}
\usepackage{booktabs}
\usepackage{array}

\newcommand{\CIC}{Contextualized Image Captioning}
\newcommand{\CCIC}{Controllable Contextualized Image Captioning}
\newcommand{\CCICshort}{Ctrl-CIC}
\newcommand{\PC}{$\mathcal{P}$-$\mathtt{Ctrl}$}
\newcommand{\RC}{$\mathcal{R}$-$\mathtt{Ctrl}$}
\newcommand{\tbf}[1]{\textcolor{blue}{\textbf{#1}}}
\newcommand{\Ext}{LongT5-Ext}
\newcommand{\Tuned}{LongT5-Tune}
\newcommand{\SentScore}{CLIPScore-Sent}

\begin{document}

\title{Controllable Contextualized Image Captioning: Directing the Visual Narrative through User-Defined Highlights} 

\titlerunning{\CCICshort{}: Directing the Visual Narrative through User-Defined Highlights}

\author{Shunqi Mao\inst{1}\orcidlink{0009-0007-0326-9004} \and
Chaoyi Zhang\inst{1}\orcidlink{0000-0001-8492-9711} 
\and
Hang Su\inst{2}\orcidlink{0009-0007-0527-9919}
\and 
Hwanjun Song\inst{3}\orcidlink{0000-0002-1105-0818}
\and 
Igor Shalyminov\inst{2}\orcidlink{0000-0001-9664-1774}
\and
Weidong Cai\inst{1}\orcidlink{0000-0003-3706-8896}}

\authorrunning{S.~Mao et al.}

\institute{The University of Sydney
\and
AWS AI Labs
\and
Korea Advanced Institute of Science and Technology
}

\maketitle
\begin{abstract}

\CIC{} (CIC) evolves traditional image captioning into a more complex domain, necessitating the ability for multimodal reasoning. 
It aims to generate image captions given specific contextual information. 
This paper further introduces a novel domain of \CCIC{} (\CCICshort{}).
Unlike CIC, which solely relies on broad context, \CCICshort{} accentuates a user-defined highlight, compelling the model to tailor captions that resonate with the highlighted aspects of the context. 
We present two approaches, Prompting-based Controller (\PC{}) and Recalibration-based Controller (\RC{}), to generate focused captions. \PC{} conditions the model generation on highlight by prepending captions with highlight-driven prefixes, whereas \RC{} tunes the model to selectively recalibrate the encoder embeddings for highlighted tokens.
Additionally, we design a GPT-4V empowered evaluator to assess the quality of the controlled captions alongside standard assessment methods.
Extensive experimental results demonstrate the efficient and effective controllability of our method, charting a new direction in achieving user-adaptive image captioning. 
Code is available at \url{https://github.com/ShunqiM/Ctrl-CIC}.

\end{abstract}    
\section{Introduction}

\begin{figure}[t]
  \centering
    \includegraphics[width=\linewidth]{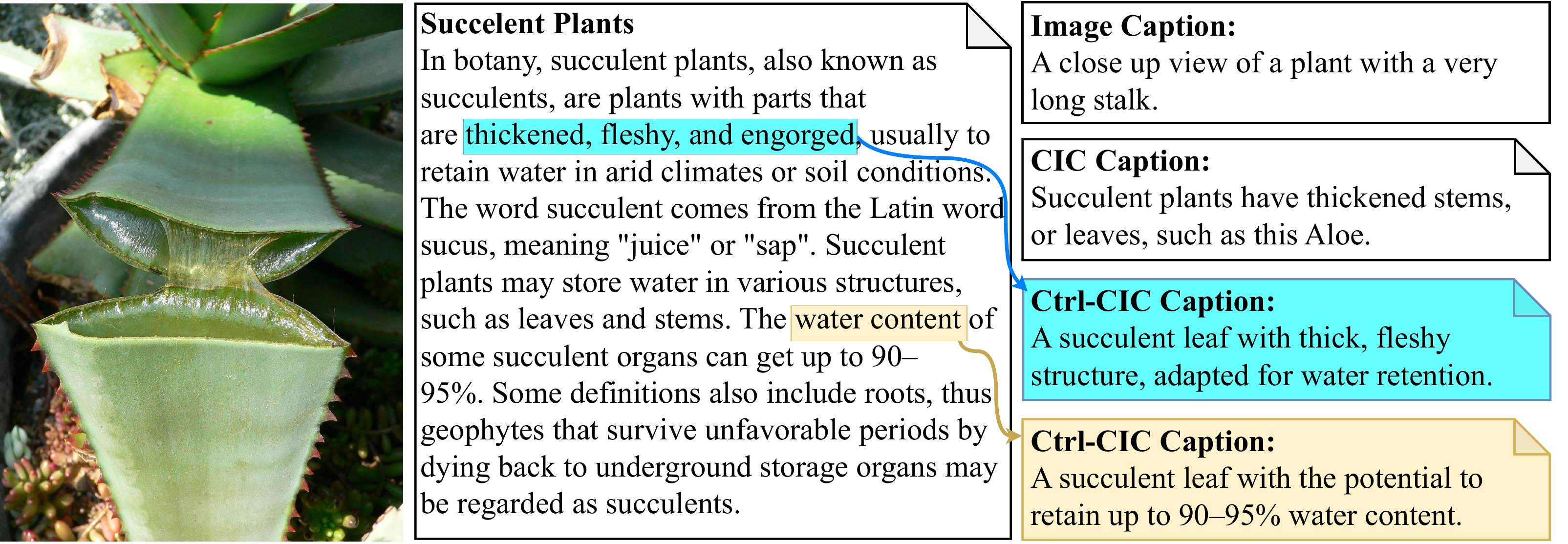}

  \caption{We introduce the \CCIC{} (\CCICshort{}) task: Given a global context, \CCICshort{} aims at generating contextualized image captions tailored to specific highlighted segments. In the presented context regarding \texttt{"succulents"}, highlights direct the caption's emphasis, underscoring distinct attributes such as its anatomical structure or water content.}
  \label{fig::teaser}
  \vspace{-3mm}
\end{figure}

Image captioning stands as a crucial task that bridges computer vision (CV) and natural language processing (NLP) domains, aiming to create sentences that effectively summarize visual content. 
Benefiting from the supervised training on extensively annotated datasets~\cite{sharma2018conceptual, plummer2015flickr30k,krause2017hierarchical}, various learning-based methods \cite{Cornia_2020_CVPR, Pan_2020_CVPR, Hu_2022_CVPR} have been developed to leverage diverse neural architectures for more accurate image captions.
The recent emergence of large language models (LLMs)~\cite{chowdhery2022palm, touvron2023llama,alayrac2022flamingo} marks an arising paradigm in the field, excelling in a variety of complex multimodal tasks under both zero-shot 
\cite{liu2023llava, su2022language, Fei_2023_ICCV} 
and few-shot \cite{NEURIPS2021_01b7575c, NEURIPS2023_e425b75b} scenarios, with image captioning as their fundamental component.
This trend highlights a growing demand for more challenging image captioning tasks to be established, which could act as evolving test-beds for benchmarking the multimodal reasoning capabilities of these cutting-edge LLM systems.

However, images are replete with rich visual cues encompassing diverse objects and varying levels of information granularity \cite{VIR}. Specifically, a single image can be interpreted and captioned in numerous ways depending on the focused visual aspects within a given context, resulting in a low inter-annotator agreement.
Consequently, generating captions for an image without considering specific context is often less controlled and ill-suited for real-world applications.
Recognizing this, Nguyen \etal~\cite{show_interpret} recently introduced the Wikipedia captioning (WikiCap) task to assist Wikipedia authors in captioning images based on the contextual backdrop of the Wikipedia page in which they appear.
Later, Burns \etal~\cite{burns2023wiki} expanded and formulated this WikiCap task into a more formal and generalized notion as \CIC{} (CIC), wherein an image is captioned along with its relevant context.

CIC has significantly enhanced the utility of image captioning models in real-world applications. However, the complexity of context presents a notable challenge: a detailed context may not yield a single ``ideal'' caption, but rather a spectrum of valid captions that resonate with different elements within the image. 
In contexts like Wikipedia articles, where topics often have multiple facets,
an image might align with multiple captions, each reflecting a unique aspect spotlighted in the associated text. Nevertheless, a mere truncation of the context to pinpoint intended highlights is not a viable solution. Such an approach, as shown in our experiments, would inadvertently omit crucial background or cues essential for the image in the context, thereby impeding the ability of the model to generate a comprehensive and precise caption.

To tackle this, we propose the Controllable Contextual Image Captioning (\CCICshort{}) task. While CIC models often deal with diverse contexts, leading to multiple plausible captions, \CCICshort{} introduces a user-controlled captioning mechanism, enabling users to pinpoint specific highlighted segments of the context that the model should prioritize, as illustrated in \cref{fig::teaser}.
By integrating the advantages of standard CIC with this novel controllability, \CCICshort{} harnesses the robust capabilities of deep models while leveraging the specificity of human intent, forging a path toward more refined and intentional captioning.

Furthermore, we present two preliminary methods for \CCICshort{}: Prompting-based Controller (\PC{}) and Recalibration-based Controller (\RC{}). These simple yet effective approaches steer models to produce captions that are relevant to specified highlights. The \PC{} method utilizes a highlight-based prefix as a prompt to guide controllable caption generation, while \RC{} recalibrates the decoder weightings so that the generated contents are augmented towards highlights. Applicable to any LMs, these methods enable controlled caption generation with finetuning.
To mitigate the lack of fine-grained highlights for \CCICshort{} task, we further propose a weakly-supervised solution that generates pseudo-highlights from the context to facilitate effective \CCICshort{} model training.

Noticeably, evaluating \CCICshort{} presents unique challenges, as it requires assessment beyond reference-based metrics such as ROUGE \cite{rouge} and BLEU \cite{bleu}, due to the absence of specific highlights and benchmarks. To this end, we leverage GPT-4 \cite{gpt4} to create test samples by extracting diverse highlights from contexts, and we adopt reference-free metrics to evaluate the alignment, relevance, and diversity of the \CCICshort{} captions. 
Moreover, we propose a GPT-4V empowered evaluator, leveraging chain-of-thought techniques for comprehensive assessment of caption quality across multiple dimensions.
Empirical results indicate that our approach is capable of generating captions that are not only diverse but also exhibit significant controllability in response to different contextual highlights.

To summarize, our key contributions are four-fold: 
\textit{(1)} We formally introduce the \CCIC{} (\CCICshort{}) problem, emphasizing the challenges of contextually influenced caption generation.
\textit{(2)} We present novel technical solutions, Prompting-based Controller (\PC{}) and Recalibration-based Controller (\RC{}), designed for the \CCICshort{} task which aligns contextual captions with user intents.
\textit{(3)} We propose an extensive evaluation pipeline, employing GPT-4(V) for highlight selection and as an additional evaluator, alongside a set of subjective measures, to comprehensively assess the controllability and overall performance of the \CCICshort{} models.
Empirical analysis demonstrates the effectiveness of the proposed controllers in \CCICshort{} task, outperforming large vision-language models with much fewer parameters.

\section{Related Work}

\subsubsection{\CIC.}
Traditional image captioning methods typically employ an encoder-decoder architecture to transform visuals into texts
\cite{cap_1, grit, cap_3}. \CIC{} (CIC) \cite{show_interpret} incorporates a textual context to the captioning task, aiming at generating captions that are not only descriptive but also contextually relevant. 
The CIC task can be traced back to news image captioning \cite{news_0}, a similar task that aims to generate descriptions for images embedded in news articles.
Biten \etal~\cite{entity_00} proposed a founding method of news image captioning by extracting named entities from context and filling them into generated caption templates.
Following works continue to explore recognizing named entities \cite{entity_2, entity_4, Tran_2020_CVPR, entity_0} or extracting factual knowledge from the context \cite{entity_1, entity_3, entity_select}, as well as large-scale pretraining \cite{context_1, show_interpret}, to facilitate context-based captioning.
The most relevant work to ours, \cite{burns2023wiki}, proposed an efficient prefix attention mechanism to handle the lengthy context, together with a large-scale CIC dataset that extends CIC to a more generalized domain instead of news articles.  
However, the context often contains redundant text or excessive information that can be equally relevant to the image. In \CCICshort{}, we guide the model with context highlights to ensure it generates highlight-specific captions.

\subsubsection{Vision-Langauge Model for Image Understanding.}
The latest state-of-the-art approaches on image captioning are dominated by finetuned large Vision-Language Models (VLMs) that bridge the gap between vision and language \cite{mplug, li2023blip2, ofa, wang2022git}. Similarly, the task of image captioning under contextual or controllable setup can be aptly addressed using VLMs. 
The majority of these models rely heavily on a vast corpus of image-text pairs for training. For example, GIT \cite{wang2022git} leverages a generative image-to-text transformer trained on 0.8B image-text pairs. Conversely, Blip2 \cite{li2023blip2} maintains static image and language models, focusing instead on training a lightweight connection module, Qformer, using 129M images. Additionally, there is a subset of VLMs
like Flamingo \cite{alayrac2022flamingo} and FROMAGe \cite{fromage} that interleave image-text sequences in a unified style. 
Beyond mere training, there is a growing interest in creating general-purpose vision-language models finetuned with instructions. InstructBlip \cite{dai2023instructblip} augments the capabilities of Blip2 by introducing instruction-aware visual feature extraction attention. Similarly, models like LLava \cite{liu2023llava} and MiniGPT-4 \cite{zhu2023minigpt} are designed for interactive conversations with multimodal data. 
However, while these models are often heavy in terms of computational resources, our approach stands out for its efficiency, requiring only $5\%$ of the parameters compared to VLMs such as LLaVA, while providing enhanced performance.

\subsubsection{Controllable Text Generation.}
Differing from standard text generation that might produce outputs based purely on training data or the input prompt, controllable text generation (CTG) allows for deriving more directed and purposeful outputs, by flexibly adjusting user-defined conditions.
A line of CTG approaches emphasizes modulating the semantics of the produced content, focusing on attributes such as sentiment \cite{sentiment_1,sentiment_2, sentiment_3, ConZIC, wang2023controllable}, topic \cite{topic_1, topic_2, topic_3}, and persona \cite{per_1, per_2, per_3}. Further research delves into lexical control during generation, for instance, integrating specific keywords or phrases \cite{keyword_1, keyword_2, zhao2024controllable}. Length-controlled language models have also been a subject of considerable investigation \cite{length_1, length_2, ding2023image}. Moreover, structured control elements, like tables and trees, have also been a topic of interest in research \cite{struct_1, struct_2}.
Traditional methods often rely on fixed categories, templates, or simple keyword rules, limiting flexibility. Our approach allows for using any contextual highlights as guidance, improving adaptability. While alignment with these highlights is crucial, we do not enforce their direct inclusion in the output.

\section{Method}


\subsection{Revisiting Contextualized Image Captioning}
Extending traditional image captioning into a domain that demands multimodal reasoning capabilities, CIC offers a more practical setup by incorporating additional contextual information into captioning. 
This context-driven method could largely benefit many real-world cases where images are often accompanied by surrounding textual content, such as news \cite{news_0} or Wikipedia captioning \cite{show_interpret}.

Formally, for a given image $ I $ and its context $ C $, where $ C $ is a set of language tokens represented as $ C = \{c_1, c_2, ..., c_n\} $, Contextual Image Captioning aims at generating captions $\mathcal{G}$ as:
\begin{equation}
\mathcal{G} = M_{\mathtt{CIC}}(I, C),
\end{equation}
where $ M_{\mathtt{CIC}}(\cdot, \cdot) $ denotes the vision-language model leveraging both the image and text. 
A CIC dataset typically contains triplet-formed data samples $D_{\mathtt{CIC}}$, which can be denoted as:
\begin{equation}
D_{\mathtt{CIC}} = \{ (C, I, T), ... \},
\end{equation}
where each triplet consists of an image $ I $, its corresponding context $ C $, and the target caption $ T $. Such datasets allow models to learn the suitable caption for images under respective contexts.

However, one intuitive challenge of CIC is that context can sometimes be overflowing with redundant text or possess excessive information, adding a layer of complexity to distinguishing the most relevant details.
Taking \cref{fig::teaser} as an example, given an image of a succulent paired with an article detailing the various characteristics of the plant, a CIC model may potentially generate a group of acceptable captions such as ``A close-up view of a succulent's stem'' or ``succulents leaves in detail'', 
each focusing on different aspects of the context.

To this end, we propose \CCICshort{} and introduce the concept of ``highlights'',  directing the model to concentrate on specific aspects of the context. 
\CCICshort{} ensures that the generated captions are more closely aligned with the human intent highlighted in these sections.
Meanwhile, the rest of the context serves primarily to furnish the model with background knowledge. 
Consequently, the model is enabled to not only generate captions that are more pertinent to the highlighted segment but also to effectively utilize the broader context's knowledge for enhanced relevance and accuracy.

\begin{figure*}[t]
  \centering
  \begin{subfigure}[t]{0.32\linewidth}
    \includegraphics[width=1\linewidth]{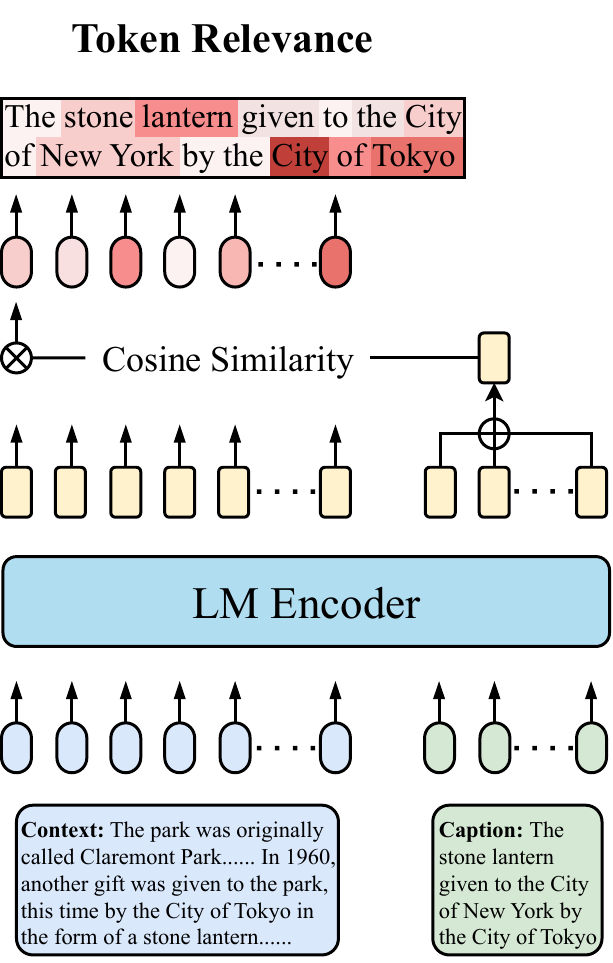}
    
    \caption{}
    \label{fig::relevance}
  \end{subfigure}
  \rule{1pt}{6.0cm} 
  \begin{subfigure}[t]{0.66\linewidth}
    \includegraphics[width=\linewidth]{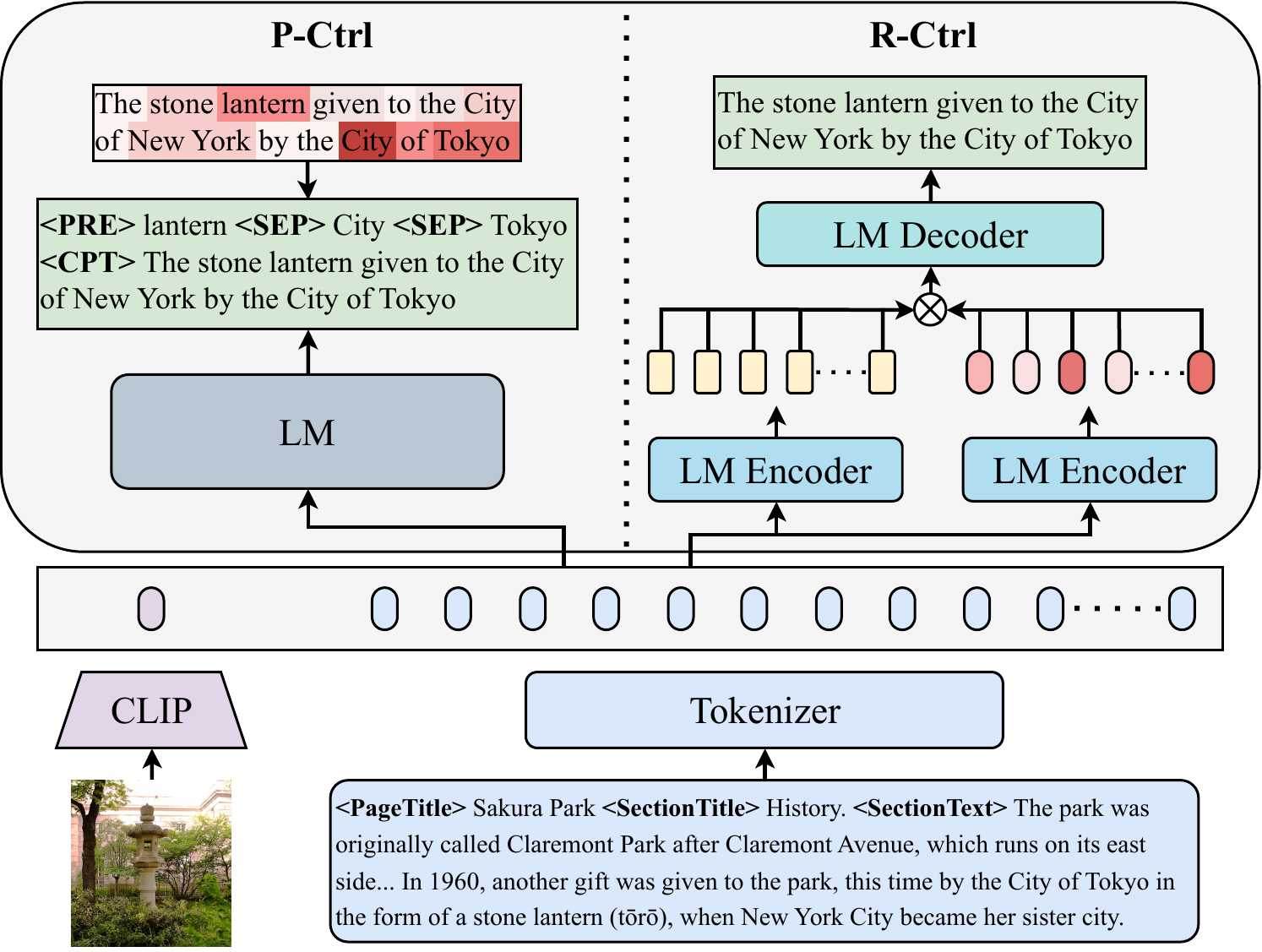}
    
    \caption{}
    \label{fig::controllers}
  \end{subfigure}

  \caption{Overview of the proposed \CCICshort{} method. 
  (a) We derive the token-level relevance scores that indicate the probability of the token being part of the highlights for the context-caption pair.
  (b) Overview of the training pipeline of the Prompting-based and Recalibration-based Controllers. For \CCICshort{} inference, the model is guided by either new prompts or recalibrated weights based on highlights, to produce controlled captions.
  }
  \label{fig::method}
\end{figure*}

\subsection{Controllable Contextual Image Captioning}

We first introduce the ``highlights'' notation $ H $, which reflects a particular intent or focus within the context. The \CCICshort{} objective now becomes:
\begin{equation}
\mathcal{G} = M_{\mathtt{CCIC}}(I, C, H).
\end{equation}
Here, the generated caption $ \mathcal{G} $ is  particularly influenced by the highlighted context $ H $, while still taking into account the overall context $ C $.
Ideally, such \CCICshort{} model should be trained on a dataset of the form:
\begin{align}
D_{\mathtt{CCIC}} = & \{(C_{i}, I_i, H_i^j, T_i^j), ...\},
\end{align}
where $i$ denotes the index of each context-image pair in the dataset, and the other index $j$ specifies the $j^{th}$ associated highlight $H_i^j$ and the target caption $T_i^j$ of the $i^{th}$ data sample. Such $H$ should satisfy the following criteria:
\begin{enumerate}
    \item The highlights, $ H = \{h_i\} $, is a collection with 
    each member $ h_i $ being a continuous subsequence of $ C $.
    
    \item Each highlighted component $ h_i $ within $ H $ could range from individual words to complete sentences, emphasizing our requirement for $ h_i $ to possess granularity at least down to the word level.

\end{enumerate}

While the target caption $T$ for traditional CIC is provided naturally in many paired image-text datasets and can be easily collected for training a standard CIC model, annotating sufficient amount of paired highlights and captions ($H_{\cdot}$, $T_{\cdot}$) for \CCICshort{} can be manpower-intensive, due to the scarcity of diverse captions corresponding to a single image and context on the Internet. 
To the best of our knowledge, we could not identify any dataset with annotations that could fit or be adapted to this description. 
Given this limitation, we construct the controllable highlights $H$ from a CIC dataset comprised of paired data $(C, I, T)$ to train \CCICshort{} models in a weakly supervised manner with the following method.

Intuitively, for a given context $C$, the highlight $H$ controlling the generation of a \CCICshort{} caption $T$ should comprise highlighted components $h$ that bear the highest relevance to the caption among other elements in $C$.
Therefore, we aim to find such $h$, as illustrated in \cref{fig::relevance}.

Firstly, the context embedding $\mathtt{Emb}_{ctx} \in R^{N_{ctx}\times 768}$ is derived through an encoder $\mathtt{Enc}(\cdot)$ as $\mathtt{Emb}_{ctx} = \mathtt{Enc}(C)$, where $N_\mathtt{ctx}$ denotes the number of tokens in context $C$.
Similarly, the embedding of the target caption $\mathtt{Emb}_{tgt} \in R^{N_{tgt}\times 768}$ can also be encoded as $\mathtt{Emb}_{tgt} = \mathtt{Enc}(T)$, followed by an average pooling among tokens to derive its global embedding $\overline{\mathtt{Emb}}_{tgt}$ and $N_{tgt}$ denotes the number of tokens in the target caption.
Next, we obtain the token-level relevance scores $S = \{s_i\}$ via computing the cosine similarity between the context and target caption as:
\begin{equation}
S = \{s_i = cos(\mathtt{Emb}_{ctx}^{i}, \, \overline{\mathtt{Emb}}_{tgt}) \mid 0 \leq i < N_{ctx}\}.
\end{equation}

The resulting token-level similarity score, $ S $, implies the relevance between each context token and the caption. They can then serve as the candidate scores for each token in the context when deciding their inclusion in the highlight. Through aggregation, these token-level relevance scores can extend seamlessly to word-level granularity with averaging: $s_{w_j} = \frac{1}{k} \sum_{l=i}^{i+k-1} s_{l}$,
where $ w_j $ denotes a single word spanning through $k$ tokens (from $ c_i $ to $ c_{i+k-1} $), and $ s_{w_j} $ indicates its word-level relevance score, thereby ensuring adaptability for highlight components $h$ of diverse forms.

The following methods are designed to ensure that the content generation process is effectively conditioned on the highlights.

\subsection{Controlled Caption Generation}

We now introduce our approaches for caption generation. To process multimodal inputs, we first employ CLIP \cite{clip} to extract the image features that are dimensionally consistent with text embeddings. Subsequently, the image feature is incorporated as a distinct token at the beginning of the text embedding sequence, allowing for the processing of multimodal inputs through a unified language model framework.
Upon finetuning, the models are trained to discern the semantic essence of the image feature, enhancing their capability to generate text outputs aligned with the given image.

Incorporating additional trainable modules to handle highlight inputs can escalate computation complexity and potentially hinder the utilization of the formidable capabilities of pretrained LLMs. We propose two simple yet effective methods: the Prompting-based Controller (\PC{}) and Recalibration-based Controller (\RC{}), enabling the model to yield controllable outputs based on highlighted context without modifying model architectures. This adaptation facilitates the easy adaptation of modern LLMs into the \CCICshort{} task by finetuning them on the controllable data. 

\subsubsection{Prompting-Based Controller}
Incorporating the Prompting-based Controller (\PC{}) allows for conditional caption generation by leveraging highlights as a prompt during decoding. First, we derive the training highlights using word-level relevance scores:
\begin{align}
H = \{ w_j \mid s_{w_j} > \theta, w_j \subseteq C \},
\end{align}
where $\theta$ is the threshold value for highlight selection. Subsequently, we assemble a prompt string with elements $w_j \in H$, delineated by the special token \mbox{\texttt{<SEP>}}. To steer the model towards conditional token generation based on the highlights, we affix this prompt string to the intended caption, as illustrated in \cref{fig::controllers}. We then finetune any modern LLM to predict these augmented texts based on the image and its context. 
The autoregressive nature of LLM enables them to condition the generation of captions based on the prefixed string, effectively learning to customize outputs in response to the given highlight prompt.
During \CCICshort{} inference, the prompt, constructed from test highlight inputs, is fed into the model as decoder inputs. This input prefix guides the model to produce captions specifically tailored to the provided highlights.

\subsubsection{Recalibration-Based Controller}
Alternatively, a \CCICshort{} model can be directly trained using token-level relevance scores with the Recalibration-based Controller (\RC{}), as depicted in \cref{fig::controllers}. 
Firstly, we normalize the cosine similarity scores to weights within unit intervals: 
\[
    \mathcal{W} = \{w_i = \frac{s_i}{2} + 0.5 \mid s_i \in \mathcal{S} \}.
\]
During training, we perform element-wise multiplication for each of the encoder token embeddings with their token weight to $e_i' = e_i * w_i$. 
Through finetuning, the model is conditioned to utilize these weights that calibrate the feature distribution for guided text generation. 
For inference purposes, we further train a weight predictor based on an additional LM encoder, the encoded embeddings produced by which are then converted to weights $\hat{e}$ within unit intervals with a linear and a sigmoid layer. 
To control caption generation, the model recalibrates the predicted weights by incrementing the token weights at $i^{th}$ position by a value of $\alpha$ if that token is highlighted, \ie, 
$\hat{e}_i' = \hat{e}_i \cdot (w_i + \alpha)$,
thereby enhancing focus on the highlighted content.

\subsubsection{Adaptability to CIC}
In addition to generating controlled captions pertinent to highlights, both proposed controllers are seamlessly adaptable for traditional CIC tasks. The \PC{} achieves this by generating captions based on a self-predicted prefix, without the need for constructing a highlights-based prompt. Meanwhile, the \RC{} accomplishes traditional CIC caption generation by adhering to the predicted weights $\hat{e}$, omitting any recalibration. The versatility of these models in performing traditional CIC tasks is further elaborated in the supplementary materials.
\section{GPT-4V based Evaluation}

As the highlights data used to train the model are constructed with a computational method rather than being curated by human expertise in real-world scenarios, we cannot entrust them as reliable references to evaluate the \CCICshort{} performance of our model. 
Instead, we established an AI-enhanced evaluation framework, capitalizing on the unparalleled proficiency of the state-of-the-art language model, GPT-4V, to assess the controllability of \CCICshort{} models.

\subsubsection{Highlights Selection}
Our \CCICshort{} models are designed to generate distinct captions for different sets of highlight inputs. To facilitate this, we employed GPT-4 to select a set of $N$ candidate highlights, consisting of words or phrases $h_1, h_2, ... h_{N}$, from each data sample.
To ensure the relevance and non-redundancy of these highlights within the context, we refined the set by removing highlights that overlap or are not found in the context.
This refinement process ensures the integrity of our test sets, which are constructed to include varying numbers of highlight segments per sample.

\begin{figure}[t!]
  \centering
    \includegraphics[width=\linewidth]{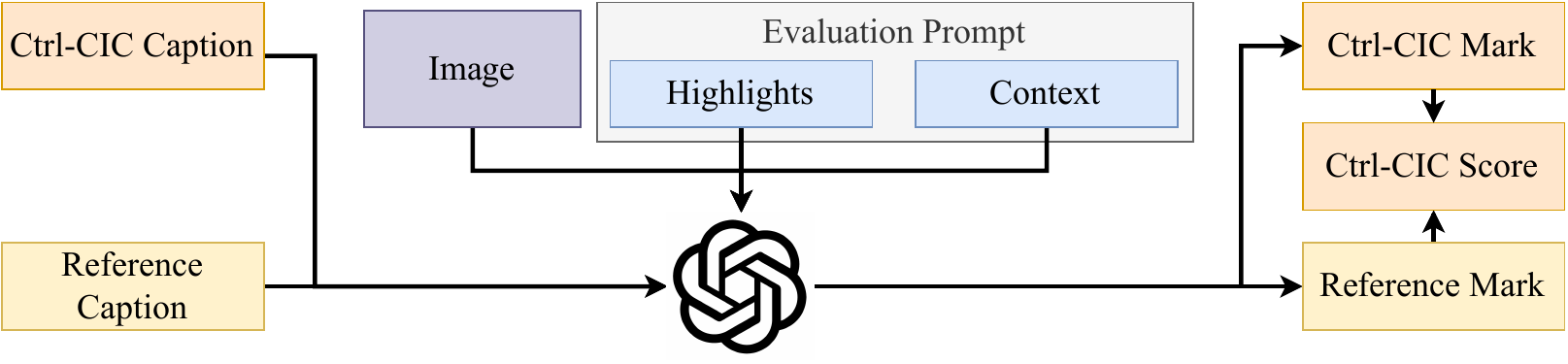}

  \caption{GPT-4V empowered evaluator for \CCICshort{} task. Given a pair of \CCICshort{} and reference captions, this GPT-4V evaluator comprehensively reasons and marks them, and derives the final score as the ratio of raw marks between \CCICshort{} and the reference caption. 
  Note: the reference caption also serves as GT for the standard CIC task.
  }
  \label{fig::eval}
\end{figure}

\subsubsection{Evaluation}
Inspired by \cite{liu2023llava}, we adopt a comparative evaluation approach to measure the quality of the \CCICshort{} captions. As illustrated in \cref{fig::eval}, we provide the GPT-4-Vision (GPT-4V) evaluator with the image, necessary context inputs, and corresponding highlights. GPT-4V then comparatively scores two captions: the reference caption (ground truth for CIC), acting as a comparative anchor, and the \CCICshort{} caption under evaluation.
These are assessed using various metrics, with scores ranging from 1 to 5.
As the ground truth for the CIC task, the reference caption inherently possesses perfect alignment with the image and its context. This characteristic makes it a suitable comparative anchor for evaluating \CCICshort{} captions.
The relative quality of a \CCICshort{} caption is then determined by dividing its mark by the reference caption mark, resulting in a score that accurately reflects the \CCICshort{} captions' quality relative to standard contextual captions.
To mitigate asymmetrical scoring bias inherent in division-based metrics, we apply a logarithmic transformation to individual scores from all test samples before computing their average. This approach symmetrically normalizes reciprocal values. Post-averaging, we exponentiate the mean score to appropriately rescale the final results, thereby preserving the relative quality assessment.

By employing this comparative scoring method, a consistent evaluative benchmark is applied to different \CCICshort{} captions. They are assessed against a common anchor, effectively reducing subjective variability in GPT-4V’s scoring. To address the positional bias inherent in the GPT-4V-based evaluators, as identified in \cite{Wang2023LargeLM}, we randomly alternate the input order of the two captions.
This approach, contrasting with individual caption scoring, ensures uniformity in evaluation and mitigates the subjective variance inherent in assessments produced by GPT-4V.

To enhance the robustness of our evaluation methodology, we utilized multi-step evaluation prompts, guiding GPT-4V to systematically score the \CCICshort{} captions in a chain-of-thought (CoT) format.
Specifically, we instruct GPT-4V to generate CoT evaluation steps following \cite{geval}, based on the predefined evaluation criteria for different metrics. 
Moreover, GPT-4V is tasked with producing analytical reasoning for each caption and comparative reasoning prior to scoring them, thereby enhancing the score quality generated through an auto-regressive process. 
Finally, we synthesize the task description, evaluation criteria, evaluation steps, and \CCICshort{} inputs to formulate the complete evaluation query. 

\section{Experiments and Results}
\subsection{Datasets and Implementation}
We conduct our experiments on Wiki-Web2M \cite{burns2023wiki}, comprising two million Wiki-pedia pages, for contextual image captioning studies. We implement our methods with the LongT5-base \cite{longt5} architecture for its balance of performance and memory efficiency, and utilize CLIP-large \cite{clip} for image feature extraction and T5-large \cite{t5} for calculating token-level relevance scores. 
We evaluate our models on the official test split provided by \cite{burns2023wiki}, with highlights sets constructed using GPT-4. We refer to this test set as Wiki-Web2M\textsubscript{full}.
The specifics of hyper-parameters and training setups are detailed in the supplementary materials.

\subsection{Baselines and Metrics}

\subsubsection{\CCICshort{} Annotations}
For better reference purposes, we alleviate the difficulty of obtaining golden standard labels for the \CCICshort{} tasks by introducing two sets of proxy annotations.
The first set comprises human-generated annotations for the traditional CIC task, referred to as CIC-GT, where these labels serve as a paradigm of optimal context relevance and image-caption consistency. 
The second set consists of \CCICshort{} labels generated by GPT-4 through a combination of context, GRIT image caption, and highlight sets. This GPT-4 reference was obtained on a subset of 5,000 randomly chosen test samples, referred to as Wiki-Web2M\textsubscript{5k}, to ensure a comprehensive evaluation.
Together, the evaluation results of these two sets of labels are indicated in gray as reference benchmarks.

\subsubsection{Baselines}
We compare the performance of our \CCICshort{} models against the following baselines.
Firstly, we adapted a pretrained LongT5 \cite{longt5} CIC model, for the \CCICshort{} task. This adaptation, named \Ext{}, selects sentences from the context based on their inclusion of the highlight, creating a highlight-centered context. 
This modified context enables the CIC model to produce captions aligned with both context and highlights, fulfilling their role as a \CCICshort{} baseline.
Additionally, we train a \Tuned{} baseline by finetuning the LongT5 model on the extracted highlight-centered context for the \CCICshort{} task. 
We also integrate the LLaVA-1.5 model \cite{liu2023llava} in our comparisons, providing a benchmark against the latest vision-language models and offering insights into the capabilities of our models relative to current state-of-the-art advancements.

\subsubsection{Metrics}
We evaluate the quality of \CCICshort{} captions using a set of reference-free metrics. We use recall (R) to evaluate the inclusion rate of the highlights in the \CCICshort{} captions. To evaluate caption diversity, we employ Div-N (D-N) \cite{divn}, measuring the proportion of distinct N-grams across five different captions generated for the same context-image pair with respect to varying highlights. In addition to the above conventional metrics, we also incorporate model-based scores. Specifically, CLIPScore (CS) \cite{clipscore} is utilized to assess the semantic alignment between the caption and the image. 
Furthermore, we introduce \SentScore{} (CS-S), a metric that calculates the cosine similarity between the CLIP text embedding of the caption and the averaged embedding of context sentences containing the highlights. Consequently, \SentScore{} reflects the extent to which the captions are relevant to the specified highlights within the context. These metrics evaluate the capability of the captions to adapt to different highlights while maintaining consistency with the image.

In terms of the GPT-4V evaluator, following \cite{geval},
prompts incorporating various metrics were used to guide GPT-4V in assessing the Ctrl-CIC captions against different criteria: \textit{Context Relevance (CR), Highlight Relevance (HR), Image Consistency (IC)}, and \textit{Overall Quality (OQ)}.
The exemplary prompts for GPT-4V-based evaluation are demonstrated in the supplementary materials.

\subsection{\CCICshort{} Results}
\begin{table}[!htb]
\caption{Evaluation on \CCICshort{} with a \textit{single} highlighted segment per sample.
Results are evaluated using conventional metrics including Recall (R), Div-1 (D-1), Div-2 (D-2), and model-based metrics including CLIPScore (CS), and \SentScore{} (CS-S).} \label{tab::ccic_single}
\centering
\scriptsize
\begin{tabular}{ll|ccccc|ccccc}
\toprule
& & \multicolumn{5}{c|}{Wiki-Web2M\textsubscript{full}} & \multicolumn{5}{c}{Wiki-Web2M\textsubscript{5k}} \\
\midrule
Model & Setting & R $\uparrow$ & D-1 $\uparrow$ & D-2 $\uparrow$ & CS $\uparrow$ & CS-S $\uparrow$& R $\uparrow$ & D-1 $\uparrow$ & D-2 $\uparrow$ & CS $\uparrow$ & CS-S $\uparrow$ \\ 
\midrule
\textcolor{gray}{CIC GT} \cite{burns2023wiki} & - & \textcolor{gray}{8.28} & \textcolor{gray}{3.24} & \textcolor{gray}{2.76}  &  \textcolor{gray}{68.66} & \textcolor{gray}{51.30}  & \textcolor{gray}{12.05} &  \textcolor{gray}{3.20} & \textcolor{gray}{2.76} &\textcolor{gray}{68.09} & \textcolor{gray}{53.00}\\
\textcolor{gray}{GPT-4} \cite{gpt4} & -  & \textcolor{gray}{-} & \textcolor{gray}{-} &  \textcolor{gray}{-} & \textcolor{gray}{-} & \textcolor{gray}{-} & \textcolor{gray}{86.10} &\textcolor{gray}{7.09} &\textcolor{gray}{8.38} & \textcolor{gray}{65.09} & \textcolor{gray}{56.33} \\
\Ext{} & Pretrained &18.45 & 5.65 & 5.59
 & 64.28 &  57.99 & 15.28 &6.23 & 6.35
 & 64.40 &  57.14  \\
\Tuned{} & Finetune  & 17.62 &6.06 &5.83  & 63.35 & 57.48 & 14.44 & 6.49 &6.45
 & 63.24 & 56.47\\
LLaVA-1.5 \cite{liu2023llava}  & Zero-shot& 36.57 &7.09 & 8.38  & \textbf{65.56} & 50.34  & 32.56 & 7.25 & 8.64  & \textbf{65.53} &47.98 \\
\midrule
\multicolumn{2}{l|}{\textit{Our Approach.}} & & & & & & & &\\
\PC{} & Finetune &57.07	& \textbf{10.23} & \textbf{10.54} & 60.91 &	\textbf{59.57} & 57.34 & \textbf{10.53} & \textbf{10.83} & 60.14 & \textbf{59.59} \\
\RC{} & Finetune  & \textbf{64.88} & 9.21 & 10.30 & 63.60 & 59.27 & \textbf{63.5} & 9.47 & 10.65 & 62.70 &  58.94\\
\bottomrule
\end{tabular}

\end{table}

\begin{table}[!htb]
\caption{Evaluation on \CCICshort{} with \textit{multiple} highlighted segments per sample. }\label{tab::ccic_multiple}
\centering
\scriptsize
\begin{tabular}{ll|ccccc|ccccc}
\toprule
& & \multicolumn{5}{c|}{Wiki-Web2M\textsubscript{full}} & \multicolumn{5}{c}{Wiki-Web2M\textsubscript{5k}} \\
\midrule
Model & Setting & R $\uparrow$ & D-1 $\uparrow$ & D-2 $\uparrow$ & CS $\uparrow$ & CS-S $\uparrow$& R $\uparrow$ & D-1 $\uparrow$ & D-2 $\uparrow$ & CS $\uparrow$ & CS-S $\uparrow$ \\ 
\midrule
\textcolor{gray}{CIC GT} \cite{burns2023wiki} & - & \textcolor{gray}{9.6} & \textcolor{gray}{3.23} & \textcolor{gray}{2.76}  &  \textcolor{gray}{67.81} & \textcolor{gray}{55.89}  & \textcolor{gray}{16.30} &  \textcolor{gray}{3.25} & \textcolor{gray}{2.78} &\textcolor{gray}{68.66} & \textcolor{gray}{55.16}\\
\textcolor{gray}{GPT-4} \cite{gpt4} & -  & \textcolor{gray}{-} & \textcolor{gray}{-} &  \textcolor{gray}{-} & \textcolor{gray}{-} & \textcolor{gray}{-} & \textcolor{gray}{84.02} &\textcolor{gray}{7.98} &\textcolor{gray}{9.82} & \textcolor{gray}{67.89} & \textcolor{gray}{53.69} \\
\Ext{} & Pretrained &13.48& 5.57 &5.53  & 64.47&60.30  & 23.94 & 5.58 &5.58 &64.97 & 61.00
  \\
\Tuned{} & Finetune  & 12.68 & 5.92 & 5.81    & 63.86 & 59.80 & 22.23 &  5.93 &5.82 & 64.01 & 60.47\\
LLaVA-1.5 \cite{liu2023llava}  & Zero-shot& 26.06 &  6.53 & 7.82 & \textbf{66.38} & 51.88  & 34.56 & 6.48 & 7.74  & \textbf{66.88} & 52.14 \\

\midrule
\multicolumn{2}{l|}{\textit{Our Approach.}} & & & & & & & &\\
\PC{} & Finetune & 44.37 & \textbf{9.45} & \textbf{10.34}  & 61.59 & 62.94 & 56.62 & \textbf{9.41} & \textbf{10.27} & 61.32 & \textbf{68.01} \\
\RC{} & Finetune  & \textbf{52.34} & 8.97 & 10.22   & 62.67 & \textbf{63.00} & \textbf{58.85} &  8.96 & 10.19 & 62.53 & 66.59\\
\bottomrule
\end{tabular}

\end{table}

\subsubsection{Quantitative Analysis}
We evaluate the performance of \CCICshort{} methods under different highlight settings. \cref{tab::ccic_single} presents the quantitative results on test samples with a singular highlighted phrase in the context, and \cref{tab::ccic_multiple} demonstrates the quantitative results when multiple highlighted phrases are present.

Under both settings, recall and \SentScore{} have indicated that both proposed models possess significantly enhanced relevance to highlights. Meanwhile, the Div-1 and Div-2 scores prove the diversity of the proposed controllers is superior to the baselines. The results indicate effective controls have been integrated into the caption generation process. 
While both Long-T5-based methods show some controllability compared to the CIC GT, their implicit highlight incorporation via extraction makes fine-grained control challenging. Additionally, truncating context to emphasize highlights often excludes vital background information, hindering the model's capacity for comprehensive captioning. 
On the other hand, our proposed controllers achieve fine-grained highlight incorporation and effectively utilize the whole context, leading to significantly stronger controllability and caption quality.

Nonetheless, the \CCICshort{} models revealed some challenges in maintaining consistency with the image content according to the CLIPScore. This observation aligns with expectations, given that the \CCICshort{} captions mainly focused on the highlighted elements, potentially leading to a narrowed or biased interpretation of the visual data. This underscores a vital direction for future enhancement.

\subsubsection{Evaluation with GPT-4V}

The GPT-4V-based evaluation results, as detailed in \cref{tab:ccic_gpt}, corroborate our findings. Notably, the \RC{} controller not only excels in `Highlight Relevance' and `Overall Quality' but also demonstrates comparable performance in `Context Relevance' to the CIC Ground Truth. This suggests that the \RC{} method effectively maintains relevance to the overall context, not just the highlighted segments.

Although the GPT-4V evaluation largely agrees with subjective assessments in aspects like highlight relevance and image consistency, some discrepancies are noted. For instance, the GPT-4V evaluator assigns LLaVA-1.5 the highest score for image consistency, surpassing even the reference labels, contrary to what CLIPScore indicates. This difference could be attributable to GPT-4V's use of contextual clues for image content assessment, as opposed to CLIPScore's reliance solely on visual information.
Despite these variances, a correlation analysis confirms that the GPT-4V-based evaluator usually aligns more closely with human judgment than the subjective metrics. Further details of this analysis are available in the supplementary materials.

\begin{table*}[!tb]
\caption{Evaluating \CCICshort{} with GPT-4V. Scores of CIC ground truth (CIC GT) that serve as the comparative anchor are always normalized to unit values. 
The comparative scores reflect relative performance; scores exceeding 1 signify superior performance compared to the CIC GT caption.
}
\label{tab:ccic_gpt}
\centering
\setlength{\tabcolsep}{10pt}
\begin{tabular}{l|cccc}
\toprule
       &      CR $\uparrow$ & HR $\uparrow$ & IC $\uparrow$ & OQ $\uparrow$     \\ 
\midrule
\textcolor{gray}{CIC GT} \cite{burns2023wiki}      &  \textcolor{gray}{1}  & \textcolor{gray}{1} & \textcolor{gray}{1}  &  \textcolor{gray}{1}   \\
\textcolor{gray}{GPT-4} \cite{gpt4} & \textcolor{gray}{1.5}    & \textcolor{gray}{2.81} & \textcolor{gray}{0.98} & \textcolor{gray}{2.00} \\ 

\Ext{} & 0.95 & 1.03 & 0.88 & 1.17 \\
\Tuned{} & 0.86    & 0.96    & 0.84   & 1.06 \\

LLaVA-1.5 \cite{liu2023llava}  & 0.98   & 1.27          & \textbf{1.05}       & 1.26 \\
\midrule
\textit{Our Approach.} \\
\PC{} & 0.94    & 1.64        & 0.75     & 1.21   \\ 

\RC{} & \textbf{1.01}    & \textbf{1.73}     & 0.71   & \textbf{1.3}  \\
\bottomrule
\end{tabular}

\vspace{-0.3cm}
\end{table*}

\subsubsection{Discussion}

While both proposed controllers exhibit commendable levels of controllability, our results indicate that the \RC{} tends to produce captions more directly related to the highlights compared to the \PC{}. Conversely, the \PC{} generates a greater diversity of captions for different highlights within the same context-image pair. This distinction likely arises because the \RC{} pushes generation towards highlight-related tokens at each decoding step, which may limit its creative scope. In contrast, the \PC{} implicitly conditions its controlled captions on the highlights implicitly, primarily through decoder attention, thus affording greater versatility in the caption generation process. Overall, our methods demonstrate superior performance compared to LLaVA-1.5 at generating controlled captions, with  a mere $5\%$ of its parameter count.

\subsubsection{Qualitative Analysis}
\begin{figure*}[t]
  \centering

  \begin{subfigure}[t]{0.49\linewidth}
    \includegraphics[width=1\linewidth]{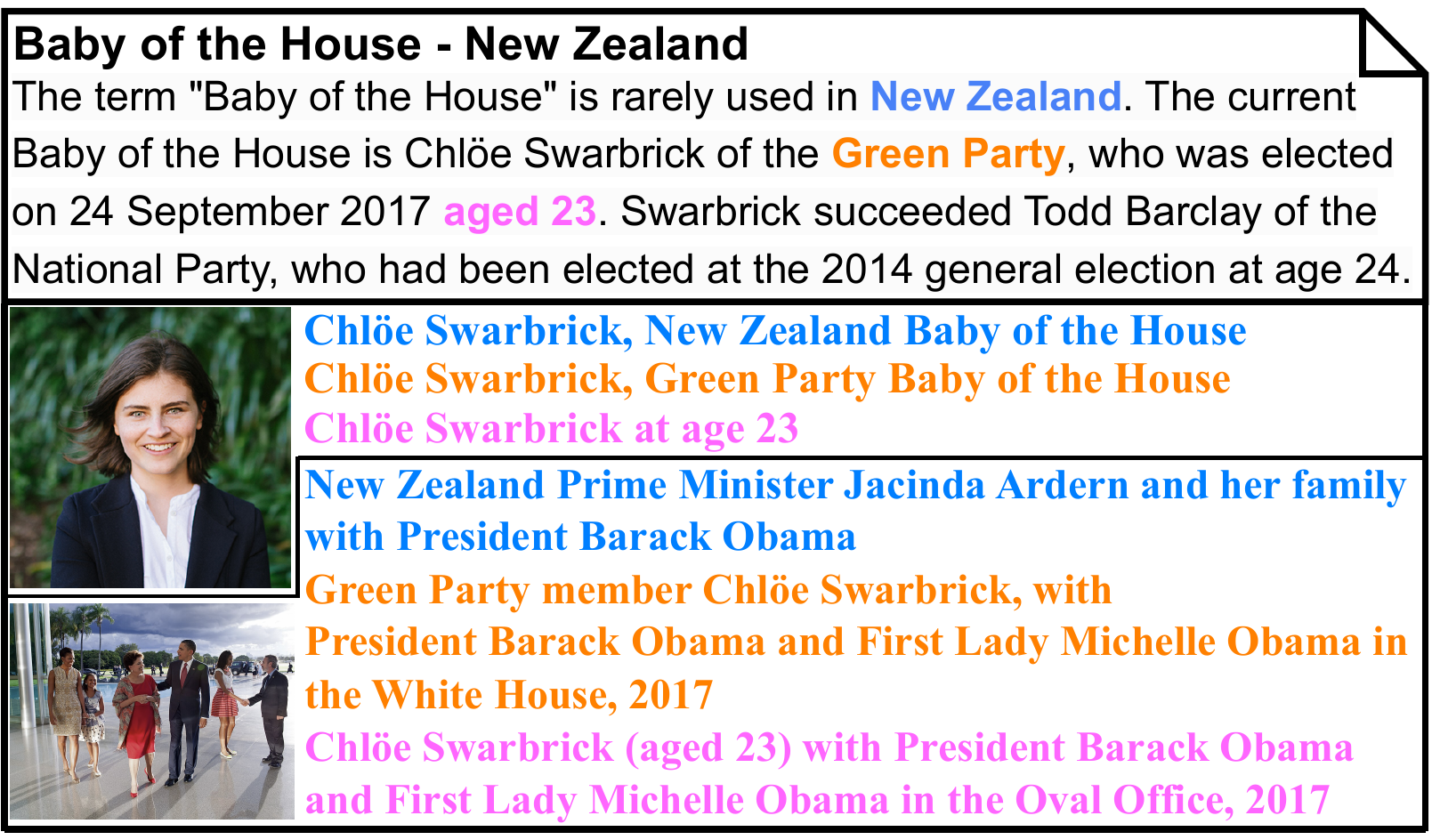}

  \end{subfigure}
  \hfill
  \begin{subfigure}[t]{0.49\linewidth}
    \includegraphics[width=\linewidth]{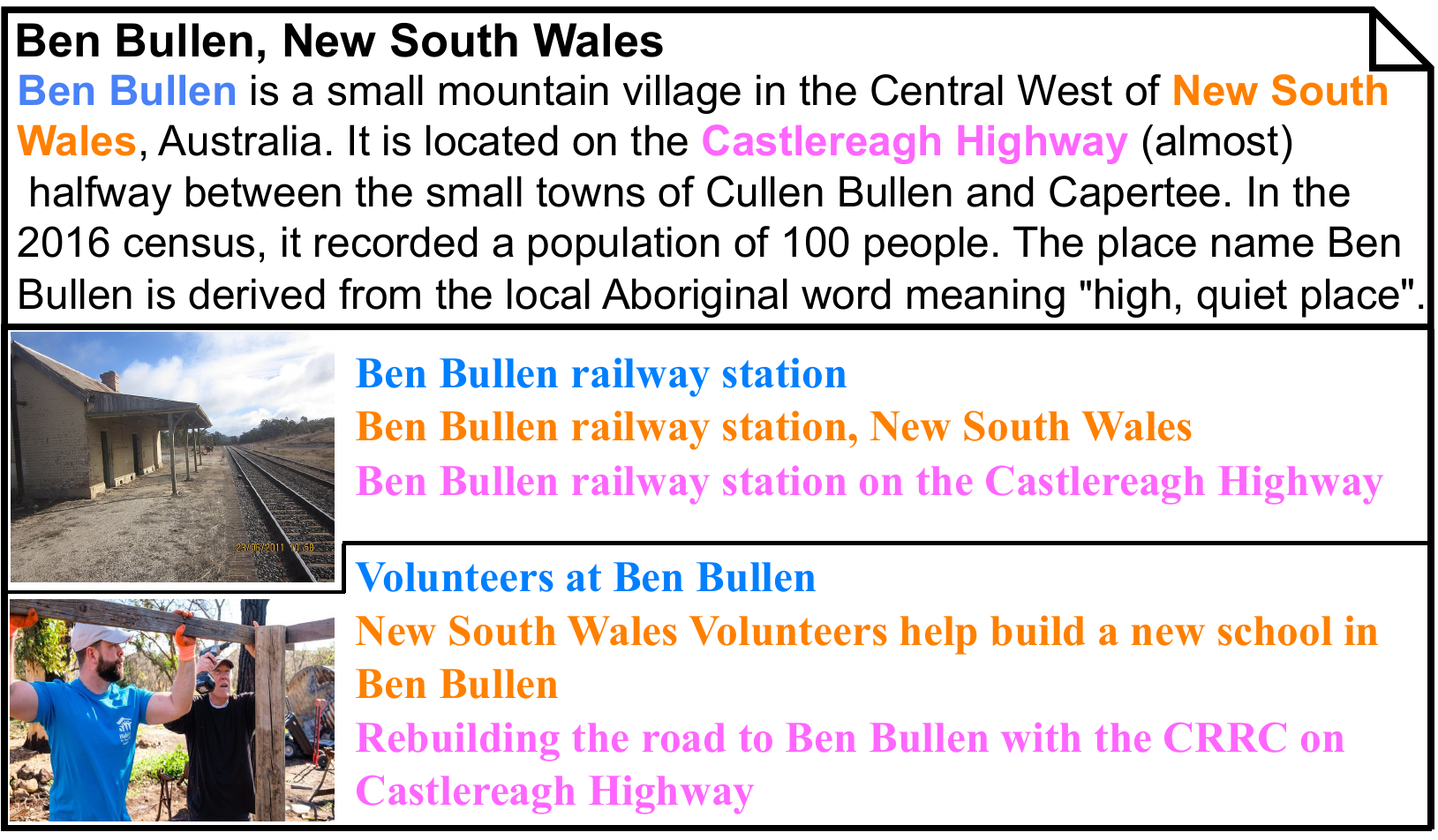}

  \end{subfigure}
\vspace{2mm}
\begin{subfigure}[b]{0.49\linewidth}
    \includegraphics[width=\linewidth]{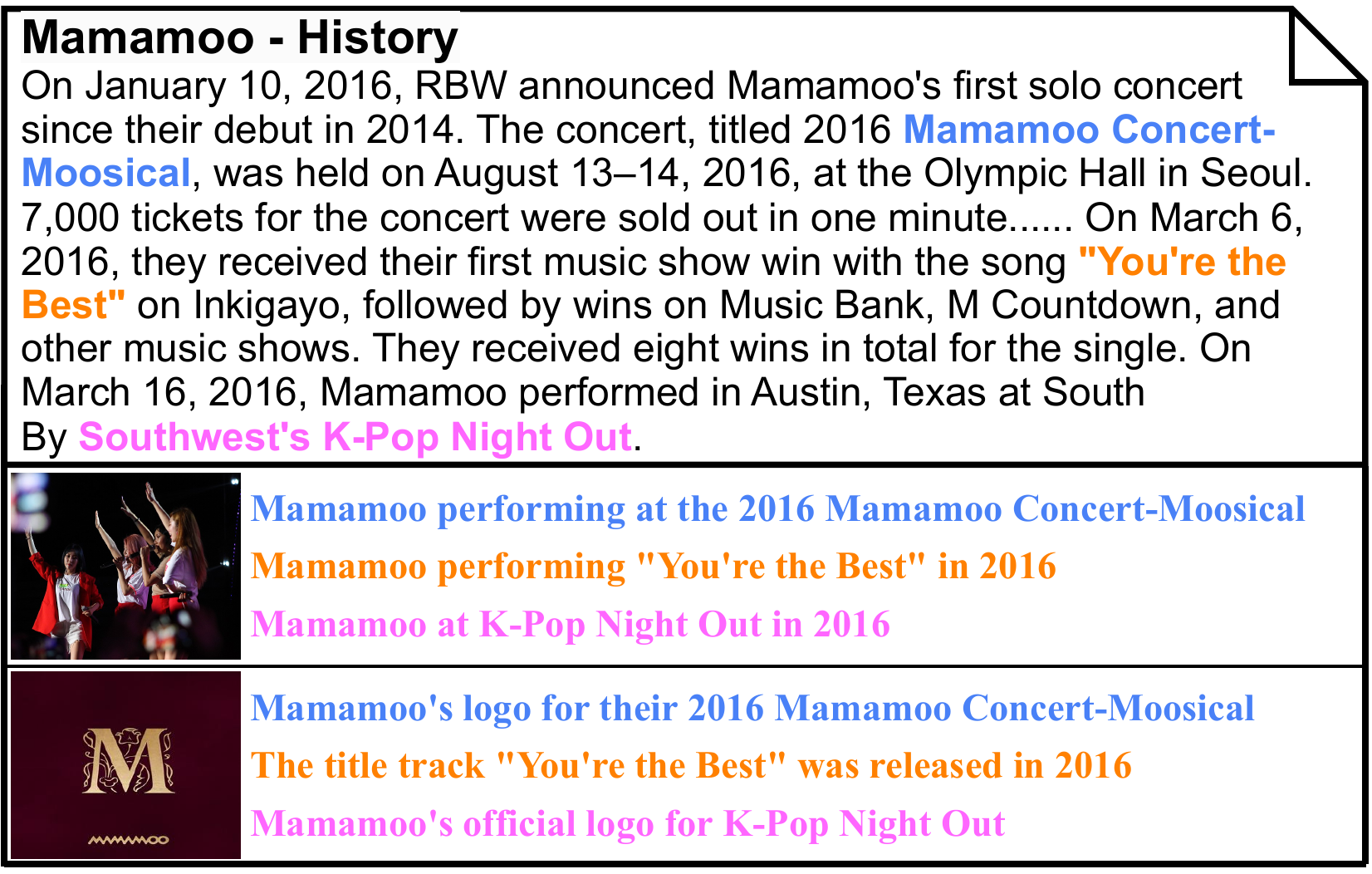}
  \end{subfigure}
  \hfill
  \begin{subfigure}[b]{0.49\linewidth}
    \includegraphics[width=1\linewidth]{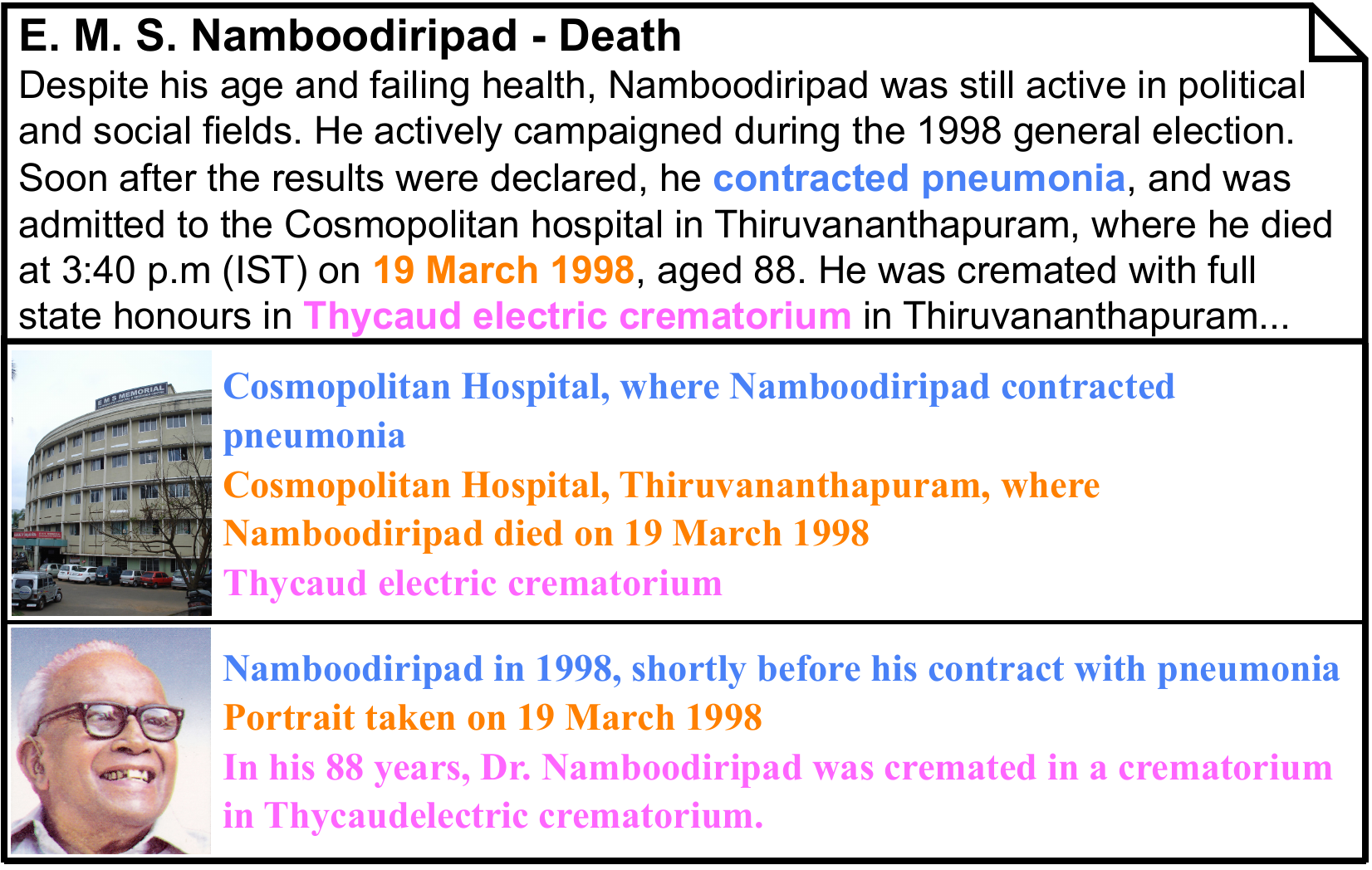}

  \end{subfigure}

  \caption{Qualitative demonstration on our \CCICshort{} results produced by \PC{}. 
  Highlights and their respective \CCICshort{} captions are aligned in colors, showing \textit{how \textbf{captions} vary with different input \textbf{images} and \textbf{highlights} for the same \textbf{context}}.
  Section titles, if any, are appended after the page title. Paragraphs in the context that are without any highlights are omitted for readability.
  }
  \label{fig::demo}
\end{figure*}

\Cref{fig::demo} shows some qualitative \CCICshort{} samples generated by the \PC{} model. It can be observed that our methods can tune the model to generate diverse captions for varied images relevant to different highlights. 
The model also exhibits a notable ability to extract relevant contextual hints to enrich the captions. For instance, in the `Mamamoo' example, the model adeptly references the correct year associated with the highlighted event. 
 While the generated captions are generally brief and feature simple linguistic structures, this reflects the style of our training data, which primarily consists of concise Wikipedia captions.
Observations reveal similar results from the \RC{} method, which are not included in the figures due to their resemblance.

\subsubsection{Limitations}
The weakly supervised training approach may limit the generalization to real-life highlights and introduce biases. Incorporating synthesized highlights in training could help mitigate this. Furthermore, the \RC{} method's uniform weight recalibration across all tokens can lead to repetitive generation of highlights. An adaptive recalibration strategy that modifies weight adjustment strength based on already generated tokens could address this issue.

\section{Conclusion}

In this work, we introduce a new challenging captioning task termed as \CCIC{} (\CCICshort{}). \CCICshort{} augments standard CIC by controlling over highlight elements, facilitating caption generation more attuned to user-specific intent. As initial solutions, we present \PC{} and \RC, two versatile approaches compatible with arbitrary LMs, enabling them to handle \CCICshort{} without compromising performance in conventional CIC tasks. We further design a GPT-4V-based evaluation framework to assess the quality of the generated \CCICshort{} captions from multiple dimensions. Empirical findings underscore the efficacy of our proposed methods, manifesting notable enhancements in controllability relative to existing captioning methods while maintaining superior efficiency.

\clearpage

%
%
\bibliographystyle{splncs04}
\bibliography{egbib}

\begin{thebibliography}{10}
\providecommand{\url}[1]{\texttt{#1}}
\providecommand{\urlprefix}{URL }
\providecommand{\doi}[1]{https://doi.org/#1}

\bibitem{alayrac2022flamingo}
Alayrac, J.B., Donahue, J., Luc, P., Miech, A., Barr, I., Hasson, Y., Lenc, K., Mensch, A., Millican, K., Reynolds, M., et~al.: Flamingo: a visual language model for few-shot learning. In: Advances in Neural Information Processing Systems (NeurIPS). vol.~35, pp. 23716--23736 (2022)

\bibitem{cap_1}
Anderson, P., He, X., Buehler, C., Teney, D., Johnson, M., Gould, S., Zhang, L.: Bottom-up and top-down attention for image captioning and visual question answering. In: Proceedings of the IEEE/CVF Conference on computer vision and pattern recognition (CVPR). pp. 6077--6086 (2018)

\bibitem{divn}
Aneja, J., Agrawal, H., Batra, D., Schwing, A.G.: Sequential latent spaces for modeling the intention during diverse image captioning. In: Proceedings of the IEEE international conference on computer vision (ICCV). pp. 4260--4269 (2019)

\bibitem{banerjee-lavie-2005-meteor}
Banerjee, S., Lavie, A.: {METEOR}: An automatic metric for {MT} evaluation with improved correlation with human judgments. In: Proceedings of the {ACL} Workshop on Intrinsic and Extrinsic Evaluation Measures for Machine Translation and/or Summarization. pp. 65--72 (2005)

\bibitem{entity_00}
Biten, A., Gomez, L., Rusinol, M., Karatzas, D.: Good news, everyone! context driven entity-aware captioning for news images. In: Proceedings of the IEEE/CVF Conference on Computer Vision and Pattern Recognition (CVPR). pp. 12458--12467 (2019)

\bibitem{burns2023wiki}
Burns, A., Srinivasan, K., Ainslie, J., Brown, G., Plummer, B.A., Saenko, K., Ni, J., Guo, M.: A suite of generative tasks for multi-level multimodal webpage understanding. In: Proceedings of the Conference on Empirical Methods in Natural Language Processing (EMNLP) (2023)

\bibitem{keyword_1}
Carlsson, F., {\"O}hman, J., Liu, F., Verlinden, S., Nivre, J., Sahlgren, M.: Fine-grained controllable text generation using non-residual prompting. In: Proceedings of the Annual Meeting of the Association for Computational Linguistics (ACL). pp. 6837--6857 (2022)

\bibitem{sentiment_1}
Chen, H., Yi, X., Sun, M., Li, W., Yang, C., Guo, Z.: Sentiment-controllable chinese poetry generation. In: Proceedings of the Twenty-Eighth International Joint Conference on Artificial Intelligence (IJCAI). pp. 4925--4931 (2019)

\bibitem{chowdhery2022palm}
Chowdhery, A., Narang, S., Devlin, J., Bosma, M., Mishra, G., Roberts, A., Barham, P., Chung, H.W., Sutton, C., Gehrmann, S., et~al.: Palm: Scaling language modeling with pathways. arXiv preprint arXiv:2204.02311  (2022)

\bibitem{Cornia_2020_CVPR}
Cornia, M., Stefanini, M., Baraldi, L., Cucchiara, R.: Meshed-memory transformer for image captioning. In: Proceedings of the IEEE/CVF Conference on Computer Vision and Pattern Recognition (CVPR) (2020)

\bibitem{dai2023instructblip}
Dai, W., Li, J., Li, D., Tiong, A.M.H., Zhao, J., Wang, W., Li, B., Fung, P., Hoi, S.: Instruct{BLIP}: Towards general-purpose vision-language models with instruction tuning. arXiv preprint arXiv:2305.06500  (2023)

\bibitem{topic_2}
Dathathri, S., Madotto, A., Lan, J., Hung, J., Frank, E., Molino, P., Yosinski, J., Liu, R.: Plug and play language models: A simple approach to controlled text generation. In: Proceedings of the International Conference on Learning Representations (ICLR) (2020)

\bibitem{ding2023image}
Ding, N., Deng, C., Tan, M., Du, Q., Ge, Z., Wu, Q.: Image captioning with controllable and adaptive length levels. IEEE Transactions on Pattern Analysis and Machine Intelligence  (2023)

\bibitem{Fei_2023_ICCV}
Fei, J., Wang, T., Zhang, J., He, Z., Wang, C., Zheng, F.: Transferable decoding with visual entities for zero-shot image captioning. In: Proceedings of the IEEE/CVF International Conference on Computer Vision (ICCV). pp. 3136--3146 (2023)

\bibitem{longt5}
Guo, M., Ainslie, J., Uthus, D., Ontanon, S., Ni, J., Sung, Y.H., Yang, Y.: {L}ong{T}5: {E}fficient text-to-text transformer for long sequences. In: North American Chapter of the Association for Computational Linguistics (NAACL). pp. 724--736 (2022)

\bibitem{VIR}
Gupta, A., Jain, R.: Visual information retrieval. Commun. ACM  \textbf{40}(5),  70–79 (1997)

\bibitem{keyword_2}
He, X.: Parallel refinements for lexically constrained text generation with {BART}. In: Proceedings of the Conference on Empirical Methods in Natural Language Processing (EMNLP). pp. 8653--8666 (2021)

\bibitem{clipscore}
Hessel, J., Holtzman, A., Forbes, M., Bras, R.L., Choi, Y.: {CLIPScore}: A reference-free evaluation metric for image captioning. arXiv preprint arXiv:2104.08718  (2022)

\bibitem{Hu_2022_CVPR}
Hu, X., Gan, Z., Wang, J., Yang, Z., Liu, Z., Lu, Y., Wang, L.: Scaling up vision-language pre-training for image captioning. In: Proceedings of the IEEE/CVF Conference on Computer Vision and Pattern Recognition (CVPR). pp. 17980--17989 (2022)

\bibitem{NEURIPS2023_e425b75b}
Huang, S., Dong, L., Wang, W., Hao, Y., Singhal, S., Ma, S., Lv, T., Cui, L., Mohammed, O.K., Patra, B., Liu, Q., Aggarwal, K., Chi, Z., Bjorck, N., Chaudhary, V., Som, S., SONG, X., Wei, F.: Language is not all you need: Aligning perception with language models. In: Oh, A., Neumann, T., Globerson, A., Saenko, K., Hardt, M., Levine, S. (eds.) Advances in Neural Information Processing Systems (NeurIPS). vol.~36, pp. 72096--72109 (2023)

\bibitem{context_1}
Kalarani, A.R., Bhattacharyya, P., Chhaya, N., Shekhar, S.: {“Let’s not Quote out of Context”}: Unified vision-language pretraining for context assisted image captioning. In: Proceedings of the Annual Meeting of the Association for Computational Linguistics (ACL) (2023)

\bibitem{topic_1}
Khalifa, M., Elsahar, H., Dymetman, M.: A distributional approach to controlled text generation. In: Proceedings of the International Conference on Learning Representations (ICLR) (2021)

\bibitem{fromage}
Koh, J.Y., Salakhutdinov, R., Fried, D.: Grounding language models to images for multimodal inputs and outputs. In: Proceedings of the International Conference on Machine Learning (ICML) (2023)

\bibitem{krause2017hierarchical}
Krause, J., Johnson, J., Krishna, R., Fei-Fei, L.: A hierarchical approach for generating descriptive image paragraphs. In: Proceedings of the IEEE conference on computer vision and pattern recognition (CVPR). pp. 317--325 (2017)

\bibitem{mplug}
Li, C., Xu, H., Tian, J., Wang, W., Yan, M., Bi, B., Ye, J., Chen, H., Xu, G., Cao, Z., Zhang, J., Huang, S., Huang, F., Zhou, J., Si, L.: m{PLUG}: Effective and efficient vision-language learning by cross-modal skip-connections. In: Proceedings of the Conference on Empirical Methods in Natural Language Processing (EMNLP). pp. 7241--7259 (2022)

\bibitem{li2023blip2}
Li, J., Li, D., Savarese, S., Hoi, S.: {BLIP-2:} bootstrapping language-image pre-training with frozen image encoders and large language models. In: Proceedings of the International Conference on Machine Learning (ICML) (2023)

\bibitem{rouge}
Lin, C.Y.: {ROUGE}: A package for automatic evaluation of summaries. In: Text Summarization Branches Out. pp. 74--81 (2004)

\bibitem{liu2023llava}
Liu, H., Li, C., Wu, Q., Lee, Y.J.: Visual instruction tuning. In: Advances in Neural Information Processing Systems (NeurIPS). vol.~36, pp. 34892--34916 (2023)

\bibitem{geval}
Liu, Y., Iter, D., Xu, Y., Wang, S., Xu, R., Zhu, C.: {G-Eval}: Nlg evaluation using gpt-4 with better human alignment. arXiv preprint arXiv:2303.16634  (2023)

\bibitem{length_1}
Liu, Y., Jia, Q., Zhu, K.: Length control in abstractive summarization by pretraining information selection. In: Proceedings of the Annual Meeting of the Association for Computational Linguistics (ACL). pp. 6885--6895 (2022)

\bibitem{adamw}
Loshchilov, I., Hutter, F.: Decoupled weight decay regularization. In: Proceedings of the International Conference on Learning Representations (ICLR) (2019)

\bibitem{length_2}
Makino, T., Iwakura, T., Takamura, H., Okumura, M.: Global optimization under length constraint for neural text summarization. In: Proceedings of the Annual Meeting of the Association for Computational Linguistics (ACL). pp. 1039--1048 (2019)

\bibitem{show_interpret}
Nguyen, K., Biten, A.F., Mafla, A., Gomez, L., Karatzas, D.: Show, interpret and tell: Entity-aware contextualised image captioning in wikipedia. In: Proceedings of the AAAI Conference on Artificial Intelligence (AAAI). pp. 1940--1948 (2022)

\bibitem{grit}
Nguyen, V.Q., Suganuma, M., Okatani, T.: {GRIT}: Faster and better image captioning transformer using dual visual features. In: Proceedings of the European Conference on Computer Vision (ECCV). pp. 167--184. Springer (2022)

\bibitem{gpt4}
OpenAI: {GPT}-4 technical report. arXiv preprint arXiv:2303.08774  (2023)

\bibitem{Pan_2020_CVPR}
Pan, Y., Yao, T., Li, Y., Mei, T.: X-linear attention networks for image captioning. In: Proceedings of the IEEE/CVF Conference on Computer Vision and Pattern Recognition (CVPR) (2020)

\bibitem{cap_3}
Pan, Y., Yao, T., Li, Y., Mei, T.: X-linear attention networks for image captioning. In: Proceedings of the IEEE/CVF Conference on Computer Vision and Pattern Recognition (CVPR) (2020)

\bibitem{bleu}
Papineni, K., Roukos, S., Ward, T., jing Zhu, W.: {BLEU}: a method for automatic evaluation of machine translation. In: Proceedings of the 40th annual meeting of the Association for Computational Linguistic (ACL). pp. 311--318 (2002)

\bibitem{plummer2015flickr30k}
Plummer, B.A., Wang, L., Cervantes, C.M., Caicedo, J.C., Hockenmaier, J., Lazebnik, S.: Flickr30k entities: Collecting region-to-phrase correspondences for richer image-to-sentence models. In: Proceedings of the IEEE international conference on computer vision (ICCV). pp. 2641--2649 (2015)

\bibitem{struct_1}
Puduppully, R., Dong, L., Lapata, M.: Data-to-text generation with content selection and planning. In: Proceedings of the AAAI Conference on Artificial Intelligence (AAAI). pp. 6908--6915 (2019)

\bibitem{entity_4}
Qu, T., Tuytelaars, T., Moens, M.F.: Visually-aware context modeling for news image captioning. arXiv preprint arXiv:2308.08325  (2023)

\bibitem{clip}
Radford, A., Kim, J.W., Hallacy, C., Ramesh, A., Goh, G., Agarwal, S., Sastry, G., Askell, A., Mishkin, P., Clark, J., et~al.: Learning transferable visual models from natural language supervision. In: Proceedings of the International Conference on Machine Learning (ICML). pp. 8748--8763 (2021)

\bibitem{t5}
Raffel, C., Shazeer, N., Roberts, A., Lee, K., Narang, S., Matena, M., Zhou, Y., Li, W., Liu, P.J.: Exploring the limits of transfer learning with a unified text-to-text transformer. Journal of Machine Learning Research  \textbf{21}(140),  1--67 (2020)

\bibitem{news_0}
Ramisa, A., Yan, F., Moreno-Noguer, F., Mikolajczyk, K.: Breakingnews: Article annotation by image and text processing. In: Proceedings of the IEEE Transactions on Pattern Analysis and Machine Intelligence (PAMI). p. 1072–1085 (2018)

\bibitem{struct_2}
Ribeiro, L.F.R., Zhang, Y., Gurevych, I.: Structural adapters in pretrained language models for {AMR}-to-{T}ext generation. In: Proceedings of the Conference on Empirical Methods in Natural Language Processing (EMNLP). pp. 4269--4282 (2021)

\bibitem{sentiment_2}
Ruan, Y., Ling, Z.: Emotion-regularized conditional variational autoencoder for emotional response generation. IEEE Transactions on Affective Computing  \textbf{14}(01),  842--848 (2023)

\bibitem{sharma2018conceptual}
Sharma, P., Ding, N., Goodman, S., Soricut, R.: Conceptual captions: A cleaned, hypernymed, image alt-text dataset for automatic image captioning. In: Proceedings of the Annual Meeting of the Association for Computational Linguistics. pp. 2556--2565 (2018)

\bibitem{per_1}
Song, H., Wang, Y., Zhang, K., Zhang, W.N., Liu, T.: {B}o{B}: {BERT} over {BERT} for training persona-based dialogue models from limited personalized data. In: Proceedings of the Annual Meeting of the Association for Computational Linguistics (ACL). pp. 167--177 (2021)

\bibitem{per_2}
Song, H., Wang, Y., Zhang, W.N., Liu, X., Liu, T.: Generate, delete and rewrite: A three-stage framework for improving persona consistency of dialogue generation. In: Proceedings of the Annual Meeting of the Association for Computational Linguistics (ACL). pp. 5821--5831 (2020)

\bibitem{sentiment_3}
Song, Z., Zheng, X., Liu, L., Xu, M., Huang, X.: Generating responses with a specific emotion in dialog. In: Proceedings of the Annual Meeting of the Association for Computational Linguistics (ACL). pp. 3685--3695 (2019)

\bibitem{su2022language}
Su, Y., Lan, T., Liu, Y., Liu, F., Yogatama, D., Wang, Y., Kong, L., Collier, N.: Language models can see: Plugging visual controls in text generation. arXiv preprint arXiv:2205.02655  (2022)

\bibitem{touvron2023llama}
Touvron, H., Martin, L., Stone, K., Albert, P., Almahairi, A., Babaei, Y., Bashlykov, N., Batra, S., Bhargava, P., Bhosale, S., et~al.: {LLaMA} 2: Open foundation and fine-tuned chat models. arXiv preprint arXiv:2307.09288  (2023)

\bibitem{Tran_2020_CVPR}
Tran, A., Mathews, A., Xie, L.: Transform and tell: Entity-aware news image captioning. In: Proceedings of the IEEE/CVF Conference on Computer Vision and Pattern Recognition (CVPR). pp. 13035--13045 (2020)

\bibitem{NEURIPS2021_01b7575c}
Tsimpoukelli, M., Menick, J.L., Cabi, S., Eslami, S.M.A., Vinyals, O., Hill, F.: Multimodal few-shot learning with frozen language models. In: Advances in Neural Information Processing Systems (NeurIPS). vol.~34, pp. 200--212 (2021)

\bibitem{cider}
Vedantam, R., Zitnick, C.L., Parikh, D.: {CIDEr}: Consensus-based image description evaluation. In: Proceedings of the IEEE/CVF Conference on Computer Vision and Pattern Recognition (CVPR). pp. 4566--4575 (2015)

\bibitem{wang2022git}
Wang, J., Yang, Z., Hu, X., Li, L., Lin, K., Gan, Z., Liu, Z., Liu, C., Wang, L.: {GIT}: A generative image-to-text transformer for vision and language. arXiv preprint arXiv:2205.14100  (2022)

\bibitem{wang2023controllable}
Wang, N., Xie, J., Wu, J., Jia, M., Li, L.: Controllable image captioning via prompting. In: Proceedings of the AAAI Conference on Artificial Intelligence (AAAI). vol.~37, pp. 2617--2625 (2023)

\bibitem{Wang2023LargeLM}
Wang, P., Li, L., Chen, L., Zhu, D., Lin, B., Cao, Y., Liu, Q., Liu, T., Sui, Z.: Large language models are not fair evaluators. arXiv preprint arXiv:2305.17926  (2023)

\bibitem{ofa}
Wang, P., Yang, A., Men, R., Lin, J., Bai, S., Li, Z., Ma, J., Zhou, C., Zhou, J., Yang, H.: Ofa: Unifying architectures, tasks, and modalities through a simple sequence-to-sequence learning framework. In: Proceedings of the International Conference on Machine Learning (ICML). pp. 23318--23340 (2022)

\bibitem{entity_3}
Wang, Y., Xu, N., Tian, H., Lv, B., Duan, Y., Li, X., Liu, A.A.: Knowledge prompt makes composed pre-trained models zero-shot news captioner. In: Proceedings of the IEEE International Conference on Multimedia and Expo (ICME). pp. 28779--2884 (2023)

\bibitem{topic_3}
Yang, P., Li, L., Luo, F., Liu, T., Sun, X.: Enhancing topic-to-essay generation with external commonsense knowledge. In: Proceedings of the Annual Meeting of the Association for Computational Linguistics (ACL). pp. 2002--2012 (2019)

\bibitem{entity_0}
Yang, X., Karaman, S., Tetreault, J., Jaimes, A.: Journalistic guidelines aware news image captioning. In: Proceedings of the Conference on Empirical Methods in Natural Language Processing (EMNLP). pp. 5162--5175 (2021)

\bibitem{ConZIC}
Zeng, Z., Zhang, H., Lu, R., Wang, D., Chen, B., Wang, Z.: {ConZIC}: Controllable zero-shot image captioning by sampling-based polishing. In: Proceedings of the IEEE/CVF Conference on Computer Vision and Pattern Recognition (CVPR). pp. 23465--23476 (2023)

\bibitem{entity_1}
Zhang, J., Fang, S., Mao, Z., Zhang, Z., Zhang, Y.: Fine-tuning with multi-modal entity prompts for news image captioning. In: Proceedings of the ACM International Conference on Multimedia (MM). p. 4365–4373. MM '22 (2022)

\bibitem{entity_2}
Zhao, W., Wu, X.: Boosting entity-aware image captioning with multi-modal knowledge graph. IEEE Transactions on Multimedia pp. 1--12 (2023)

\bibitem{zhao2024controllable}
Zhao, Y., Liu, Y., Guo, Z., Wu, W., Gong, C., Wan, F., Ye, Q.: Controllable dense captioner with multimodal embedding bridging. arXiv preprint arXiv:2401.17910  (2024)

\bibitem{per_3}
Zhong, P., Zhang, C., Wang, H., Liu, Y., Miao, C.: Towards persona-based empathetic conversational models. In: Proceedings of the Conference on Empirical Methods in Natural Language Processing (EMNLP). pp. 6556--6566 (2020)

\bibitem{entity_select}
Zhou, M., Luo, G., Rohrbach, A., Yu, Z.: Focus! relevant and sufficient context selection for news image captioning. In: Findings of the Association for Computational Linguistics: Conference on Empirical Methods in Natural Language Processing (EMNLP). pp. 6078--6088 (2022)

\bibitem{zhu2023minigpt}
Zhu, D., Chen, J., Shen, X., Li, X., Elhoseiny, M.: {MiniGPT-4}: Enhancing vision-language understanding with advanced large language models. arXiv preprint arXiv:2304.10592  (2023)

\end{thebibliography}

\clearpage
\setcounter{page}{1}
\newpage
{
\centering
\Large
\textbf{Controllable Contextualized Image Captioning: Directing the Visual Narrative through User-Defined Highlights}\\
\vspace{0.5em}Supplementary Material \\
\vspace{1em}
}

\section{Datasets and Implementation Details}\label{sec::dataset}
\normalsize

\subsubsection{Dataset Details.} We conduct our experiments on Wiki-Web2M \cite{burns2023wiki}, a large-scale webpage dataset containing two million Wikipedia article pages with text and images organized by sections, allowing context to be defined easily. Specifically, we follow the official split on the contextual image captioning task, with 2,222,814/124,703/124,188 samples for train/val/test split respectively. Each sample consists of an image paired with its relevant context derived from the webpage data. The structured data allows for a more flexible definition of context at different levels. In the presented experiment, the context of images is constructed by combining the page title, section title, section text, and captions of the remaining images in the section of the webpage.

\subsubsection{Model Details.}
We adopt LongT5 (tglobal-base) \cite{longt5} as the underlying architecture for both proposed methods as well as the weight predictor for \RC{}, due to its superior efficiency compared to Prefix Global used in \cite{burns2023wiki} with only a marginal trade-off in performance. 
We use CLIP-large \cite{clip} to extract the frozen image feature, and we adopt T5-large \cite{t5} to derive the token-level relevance scores for training.

\subsubsection{Hyperparameter Details.}
Input texts are truncated to a length of $512$ tokens, and the output length is set to $128$ for all methods except \PC{}, where an output length of $192$ is used in account for the extra prompting tokens. For \PC{}, the threshold value to derive the training highlights prompt is set to $\theta = 0.3$. 
Furthermore, we impose an upper limit of 40 on the number of prompting words during training to prevent scenarios where an excessively lengthy prompt could constrain the space available for the caption.
For \RC{}, we set the weight recalibration value to $\alpha = 0.1$ for controlled caption generation. 
During training, all methods are optimized with an AdamW \cite{adamw} optimizer with a linearly decaying learning rate of $5e-5$ and betas $(0.9, 0.999)$. The models are trained on a single NVIDIA RTX 3090 GPU, with a batch size of $12$ for 3 million steps. In addition, the weight predictor for \RC{} is trained using cross entropy loss with the same optimizer and a batch size of $20$ for 1 million steps.

\subsubsection{GPT-4V Evaluation Details.}
In terms of highlights selection, we prompt GPT-4 to select $N = 10$ highlighted component candidates from contexts, and construct the evaluation sets with different numbers of highlights based on the filtered highlights. Contexts with insufficient numbers of selected highlights are also filtered during constructing \textit{single} and \textit{multiple} highlights test sets.
For GPT-4V empowered evaluation, we randomly select 50 samples from the test set, on which the GPT-4V is prompted to evaluate the \CCICshort{} captions from different methods.

\section{Model Cost Analysis}

\begin{table*}[!tb]
\caption{Cost analysis for \CCICshort{} models. The per-sample inference time are estimated based on the average query expenses. The \Tuned{} method, having identical architecture and inputs as \Ext{}, is omitted from the table. The parameter count for GPT-4, unofficially reported as 8 $\times$ 220 billion, remains unconfirmed. Notably, besides the vast parameter size and time cost, GPT-4 also poses a monetary cost of approximately \$0.0059 per \CCICshort{} query.
}
\label{tab:cost}
\centering
\begin{tabular}{l|>{\centering\arraybackslash}p{30mm} |>{\centering\arraybackslash}p{30mm}}
\toprule
\multicolumn{1}{l|}{Model} & \multicolumn{1}{c|}{Parameters (B)} & \multicolumn{1}{c}{Time (s)} \\
\midrule
GPT-4 & 8 $\times$ 220 \textsuperscript{\dag}  & 0.563 \\
\Ext{} & 0.248 & 0.169 \\
LLaVA-1.5 & 7.063 & 0.634  \\
\midrule
\PC{} & 0.248 & 0.166 \\
\RC{} & 0.357 & 0.223 \\

\bottomrule
\end{tabular}
\end{table*}

\begin{table*}[!tb]
\caption{Correlation analysis between human judgment and different metrics. The `G-' prefixed rows represent the results from GPT-4V evaluation, whereas the `H-' prefixed column headings denote human assessments. `CS' and `CS-S' stand for CLIPScore and \SentScore{} respectively. The metric owning best human alignment, is highlighted in \textbf{bold}, and the second-best one is \underline{underlined}.
}

\label{tab:ccic_corr}
\centering
\scriptsize
\begin{tabular}{l|>{\centering\arraybackslash}p{8mm} >{\centering\arraybackslash}p{8mm} >{\centering\arraybackslash}p{8mm}|>{\centering\arraybackslash}p{8mm} >{\centering\arraybackslash}p{8mm} >{\centering\arraybackslash}p{8mm}|>{\centering\arraybackslash}p{8mm} >{\centering\arraybackslash}p{8mm} >{\centering\arraybackslash}p{8mm}|>{\centering\arraybackslash}p{8mm} >{\centering\arraybackslash}p{8mm} >{\centering\arraybackslash}p{8mm}}
\toprule
 & \multicolumn{3}{c|}{H-CR} & \multicolumn{3}{c|}{H-HR} & \multicolumn{3}{c|}{H-IC} & \multicolumn{3}{c}{H-OQ} \\
 & $r$ &  $\rho$ & $\tau$ &  $r$ & $\rho$ & $\tau$ &  $r$ & $\rho$ & $\tau$ &  $r$ & $\rho$ & $\tau$ \\
\midrule
CS & 0.479 & 0.476 & 0.344 & -0.198 & -0.228 & -0.171 & 0.546 & 0.571 & 0.438 & 0.231 & 0.198 & 0.156 \\
CS-S & 0.491 & 0.511 & 0.375 & 0.241 & \underline{0.323} & 0.233 & 0.336 & 0.183 & 0.063 & 0.501 & 0.614 & 0.469 \\
Recall & -0.032 & 0.129 & 0.112 & \textbf{0.401} & 0.309 & \underline{0.267} & -0.200 & -0.207 & -0.180 & 0.131 & 0.258 & 0.225 \\
\midrule
G-CR & \textbf{0.665} & \textbf{0.791} & \textbf{0.639} & 0.217 & 0.232 & 0.150 & 0.547 & 0.747 & \underline{0.589} & 0.570 & 0.573 & 0.488 \\
G-HR & 0.476 & 0.691 & 0.560 & \underline{0.381} & \textbf{0.495} & \textbf{0.317} & 0.277 & 0.351 & 0.336 & \underline{0.572} & \textbf{0.734} & \textbf{0.560} \\
G-IC & \underline{0.629} & 0.694 & 0.548 & -0.001 & 0.012 & 0.032 & \textbf{0.690} & \textbf{0.778} & 0.581 & 0.410 & 0.382 & 0.323 \\
G-OQ & 0.613 & \underline{0.760} & \underline{0.606} & 0.224 & 0.264 & 0.163 & \underline{0.594} & \underline{0.769} & \textbf{0.623} & \textbf{0.594} & \underline{0.626} & \underline{0.525} \\
\bottomrule
\end{tabular}
\end{table*}

\Cref{tab:cost} presents the cost analysis in terms of parameter size and inference time for \CCICshort{} models. 
For fair comparisons, input contexts across all methods are uniformly truncated to the same length.
With a common LongT5 backbone architecture, both controller methods enhance controllability relative to \Ext{}, while maintaining similar levels of efficiency. \RC{} exhibits marginally lower efficiency than \PC{} due to its additional weight predictor. Nonetheless, both controllers substantially outperform GPT-4 and LLaVA-1.5 in terms of memory and computational efficiency. Notably, while GPT-4 generates high-quality \CCICshort{} captions fairly efficiently, its high monetary costs limit its practicality in various real-world scenarios. In conclusion, our methods offer more efficient and cost-effective solutions for the \CCICshort{} task.

\section{Correlation Analysis of GPT-4V Evaluation}
\begin{figure*}[!tb]
    \centering
    \includegraphics[width=.9\linewidth]{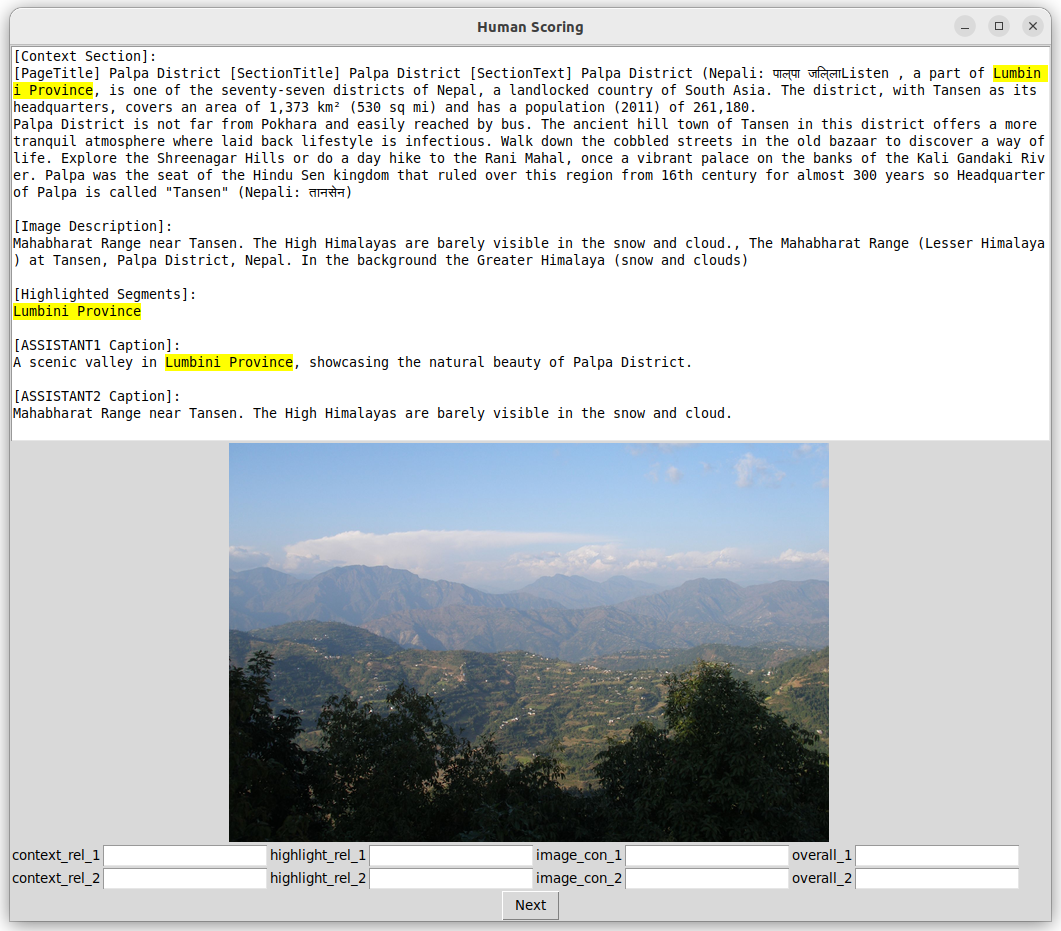}
    \caption{UI of the scoring application for human evaluation. The top text box displays the context, image description, and the highlighted segments, along with the two captions to be scored based on Context Relevance, Highlight Relevance, Image Consistency, and Overall Quality. Human markers will review the context and score both the reference and \CCICshort{} captions regarding each metric respectively.}
    \label{fig:ui}
\end{figure*}
We conduct human evaluations on \CCICshort{} captions to validate the alignment between the GPT-4V empowered evaluator and human judgment. using the same four metrics: 
\textit{Context Relevance (CR), Highlight Relevance (HR), Image Consistency (IC)}, and \textit{Overall Quality (OQ)}.
As depicted in \cref{fig:ui}, we present the contextual information and anonymous captions to human markers and ask them to score the captions based on each metric. The human scoring of the \CCICshort{} caption is then calculated based on the relative ratio of its raw mark to the reference caption's mark.

We conduct correlation analysis to examine the agreement between human judgment and metrics used in the paper. 
As shown in \cref{tab:ccic_corr}, Pearson ($r$), Spearman ($\rho$), and Kendall's tau ($\tau$) coefficients are utilized to determine the strength of association between human judgment and metric scores. 
Div-N are not included in the analysis as they aim to evaluate the diversity between a set of captions on the same context-image pair, and are irrelevant to qualitative aspects as measured by other metrics.

It is notable that the metrics generated by the GPT-4V empowered evaluator frequently exhibit the highest correlations when compared to their human-judged counterparts, as evidenced by the corresponding pairs in the analysis (\eg, `G-CR' compared to `H-CR'). 
For example, in assessing image-text consistency, `G-IC' shows a higher correlation with `H-IC' compared to CLIPScore. This higher correlation with human judgment may stem from GPT-4V's ability to utilize contextual clues in evaluating image content, in contrast to CLIPScore, which relies exclusively on visual information.

We also find that `G-HR' shows correlation levels with `H-OQ' that are akin to those observed for `G-OQ'. This is attributed to the different evaluative focuses: human annotators value highlight relevance more in their assessments, while the GPT-4V empowered evaluator prioritizes context relevance and image consistency.
This observation is further corroborated by examining the correlation between `G-OQ' and human judgment across various aspects, where it is noted that `H-HR' has the least correlation with `G-OQ'.

Furthermore, the correlation analysis reveals an intriguing challenge in \CCICshort{} captions: achieving a close relationship with highlights while also maintaining high consistency with the image content. Metrics that evaluate the relevance of captions to highlights often show zero or negative correlations with those assessing caption-image consistency. This finding supports our observations on the \CCICshort{} evaluation results, where the proposed controllers, being more focused on highlights, tend to exhibit lower image consistency.

\section{Adaptability of \CCICshort{} Models to Standard CIC Task} 
In this subsection, we explore the adaptability and robustness of our \CCICshort{} models when applied to the traditional CIC task. As previously noted, \PC{} is capable of autonomously generating CIC captions based on the prefix prompts generated during the auto-regressive decoding, while our \RC{} can be effectively adapted to the CIC task by maintaining the predicted weights without any recalibration. This section presents the experimental results of these adaptations, showcasing the flexibility and efficacy of our models in transitioning between \CCICshort{} and CIC tasks.

\begin{table*}
\centering
\caption{Evaluation on standard CIC task on the test split of Wiki-Web2M
benchmark (Wiki-Web2M\textsubscript{test}). GPT-4 CIC results are obtained and reported on a 1,000 sample subset (Wiki-Web2M\textsubscript{1k}). 
* denotes an extra input token length of 1,024 used. 
GPT-4 ($\cdot$) denotes the zero-shot inference of GPT-4 with different inputs.
}
\label{tab:cic_main}
\begin{tabular}{ll|cccc|cccc}
\toprule
& & \multicolumn{4}{c|}{Wiki-Web2M\textsubscript{test}} & \multicolumn{4}{c}{Wiki-Web2M\textsubscript{1k}} \\
\midrule
Model & Setting & B-4 $\uparrow$ & R-L $\uparrow$ & C $\uparrow$ & M $\uparrow$ & B-4 $\uparrow$ & R-L $\uparrow$ & C $\uparrow$ & M $\uparrow$ \\ 
\midrule
Prefix Global* \cite{burns2023wiki} & Finetune & 10.92 & 36.21 & 148.53 & - & - & - & - & - \\
LongT5 \cite{longt5} & Finetune & 10.31 & 36.00 & 142.77 & 30.58 & 9.42 & 35.01 & 128.03 & 29.88 \\

InstructBlip \cite{dai2023instructblip} & Zero-shot & 1.40 & 14.03 & 12.91 & 16.36 & 1.39 & 13.70 & 14.97 & 15.93 \\
LLaVA-1.5 \cite{liu2023llava}  & Zero-shot & 2.55& 16.48& 25.07 & 16.81& 2.71 &16.06 & 23.63 & 16.40\\
GPT-4 \cite{gpt4} (w/ GRIT) & Zero-shot & - & - & - & - & 4.81 & 21.55 & 40.05 & 23.13 \\
GPT-4 \cite{gpt4} (w/ Attri) & Zero-shot & - & - & - & - & 9.29 & 32.54 & 65.52 & 40.48 \\
\midrule
\textit{Our Approach.}  & & & & & & & & &\\
\PC{} & Finetune & 10.88 & 34.48 & 130.80 & 30.78 & 10.24 & 33.37 & 119.94 & 29.98 \\
\RC{} & Finetune & 10.31 & 34.58 & 130.12 & 30.83 & 10.08 & 33.15 & 117.56 & 29.74 \\
\bottomrule
\end{tabular}

\end{table*}

\subsubsection{Metrics.}
Following Wiki-Web2M \cite{burns2023wiki}, we report BLEU-4 (B-4) \cite{bleu} , RougeL (R-L) \cite{rouge}, CIDEr (C) \cite{cider}. we also compute Meteor (M) \cite{banerjee-lavie-2005-meteor} for CIC evaluation.

\subsubsection{Baselines.}
Our primary baseline of comparison is the Prefix Global method \cite{burns2023wiki}, which stands as the sole existing CIC benchmark on the Wiki-Web2M dataset \cite{burns2023wiki}. Owing to computational constraints, we reference results directly from \cite{burns2023wiki}. For a more computation-friendly benchmark, we adapted the \cite{burns2023wiki} baseline employing LongT5 \cite{longt5}. 
Similar to \CCICshort{}, we benchmark the zero-shot ability of Instruct-Blip \cite{li2023blip2}, LLaVA-1.5 \cite{liu2023llava}, and GPT-4 \cite{gpt4} on the CIC task.  
For GPT-4, we report results on a subset of 1,000 samples, referred to as Wiki-Web2M\textsubscript{1k}, with two types of text-based image descriptions: image captions produced by GRIT \cite{grit}, and the image attribution text from the dataset.

\subsubsection{Results.}
The performance of models on the CIC task is shown in \cref{tab:cic_main}. Compared to the LongT5 baseline,  both \PC{} and \RC{} exhibit higher or comparable scores in B-4 and Meteor metrics, 
indicating our proposed approaches are effective in generating high-quality contextual captions for the CIC task.  
Although our models do not reach the performance levels of the Prefix Global method, this is understandable given the computational constraints that limit both our models and the LongT5 baseline to shorter input lengths. The Prefix Global architecture has been shown to be more effective but also more computationally intensive than the LongT5 architecture \cite{burns2023wiki}. In contrast, zero-shot models demonstrate less competitiveness in the CIC task. InstructBlip, for instance, struggles to generate coherent captions due to its limitations in processing extensive context. Other CIC captions generated by zero-shot models show notable improvement over InstructBlip, producing decent captions but still not reaching the accuracy level of fine-tuned models. The evaluation results demonstrate the seamless adaptability of the proposed \CCICshort{} controllers to the standard CIC task, with minimal sacrifice on the performance.

\section{Analysis on the Importance of Visual Inputs}
Some images in \CCICshort{} simply depict the primary object in context, implying that textual context alone might suffice for generating adequate \CCICshort{} captions and downplay the importance of visual inputs. However, it is more commonly observed that images contribute essential supplementary information that enhances understanding beyond mere illustration of the context, underscoring the vital role of visual inputs in the \CCICshort{} tasks.
Quantitatively, to demonstrate the significance of visual inputs, we provide a zero-shot evaluation using Llama-2 as a pure language model (LLM) baseline without any visual or image description inputs, as detailed in \cref{tab::llama,tab::llama-geval}. Observations show that employing an LM-based model that relies solely on textual context for \CCICshort{}, while enhancing relevance with highlights and context, reduces image consistency and overall caption quality.

\begin{table}[!tb]
\caption{Llama-2 quantitative results in a pure textual-context-based \CCICshort{} formulation.} \label{tab::llama}
\centering
\scriptsize
\begin{tabular}{ll|ccccc|ccccc}
\toprule
& & \multicolumn{5}{c|}{Wiki-Web2M\textsubscript{full}} & \multicolumn{5}{c}{Wiki-Web2M\textsubscript{5k}} \\
\midrule
Model & Split & R $\uparrow$ & D-1 $\uparrow$ & D-2 $\uparrow$ & CS $\uparrow$ & CS-S $\uparrow$& R $\uparrow$ & D-1 $\uparrow$ & D-2 $\uparrow$ & CS $\uparrow$ & CS-S $\uparrow$ \\ 
\midrule
\textcolor{gray}{CIC GT} \cite{burns2023wiki} & Single & \textcolor{gray}{8.28} & \textcolor{gray}{3.24} & \textcolor{gray}{2.76}  &  \textcolor{gray}{68.66} & \textcolor{gray}{51.30}  & \textcolor{gray}{12.05} &  \textcolor{gray}{3.20} & \textcolor{gray}{2.76} &\textcolor{gray}{68.09} & \textcolor{gray}{53.00}\\

Llama-2 \cite{touvron2023llama}  & Single & 65.26 & 10.11 & 12.35 & 53.12 & 65.97 & 61.36 & 10.15 & 12.42 & 52.63 & 65.93 \\
\midrule
\textcolor{gray}{CIC GT} \cite{burns2023wiki} & Multiple & \textcolor{gray}{9.6} & \textcolor{gray}{3.23} & \textcolor{gray}{2.76}  &  \textcolor{gray}{67.81} & \textcolor{gray}{55.89}  & \textcolor{gray}{16.30} &  \textcolor{gray}{3.25} & \textcolor{gray}{2.78} &\textcolor{gray}{68.66} & \textcolor{gray}{55.16}\\
Llama-2 \cite{touvron2023llama}  & Multiple & 46.25 & 9.77 & 11.96 & 51.96 & 68.06 & 46.30 & 9.88 & 11.92 & 51.68 & 68.30 \\
\bottomrule
\end{tabular}
\end{table}

\begin{table*}[!t]
\caption{Llama-2 evaluation results with GPT-4V.
}
\label{tab::llama-geval}
\centering
\setlength{\tabcolsep}{10pt}
\begin{tabular}{l|cccc}
\toprule
       &      CR $\uparrow$ & HR $\uparrow$ & IC $\uparrow$ & OQ $\uparrow$     \\ 
\midrule
\textcolor{gray}{CIC GT} \cite{burns2023wiki}      &  \textcolor{gray}{1}  & \textcolor{gray}{1} & \textcolor{gray}{1}  &  \textcolor{gray}{1}   \\
 Llama-2  \cite{touvron2023llama} & 1.28 & 2.08 & 0.61 & 1.10 \\
\bottomrule
\end{tabular}

\end{table*}

\section{Qualitative Analysis on \CCICshort{}}
\begin{figure*}[!htb]
\vspace{-0.2cm}
    \centering
    \includegraphics[width=0.9\linewidth]{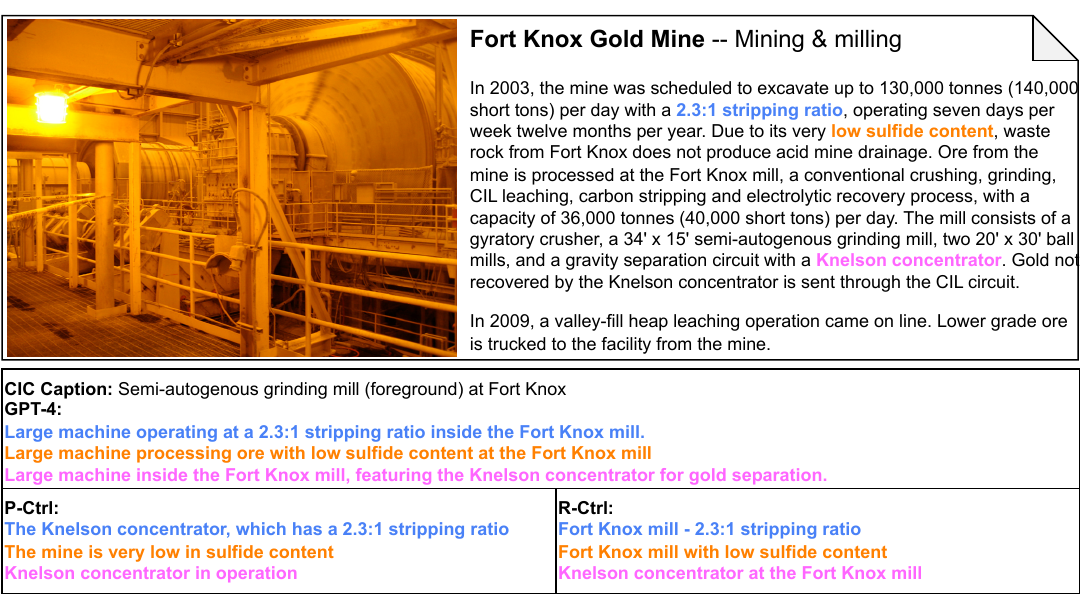}
    \caption{\CCICshort{} Captions Sample A}
    \label{fig:s1}
    \vspace{-0.15cm}
\end{figure*}

\begin{figure*}[!htb]
    \centering
    \includegraphics[width=0.9\linewidth]{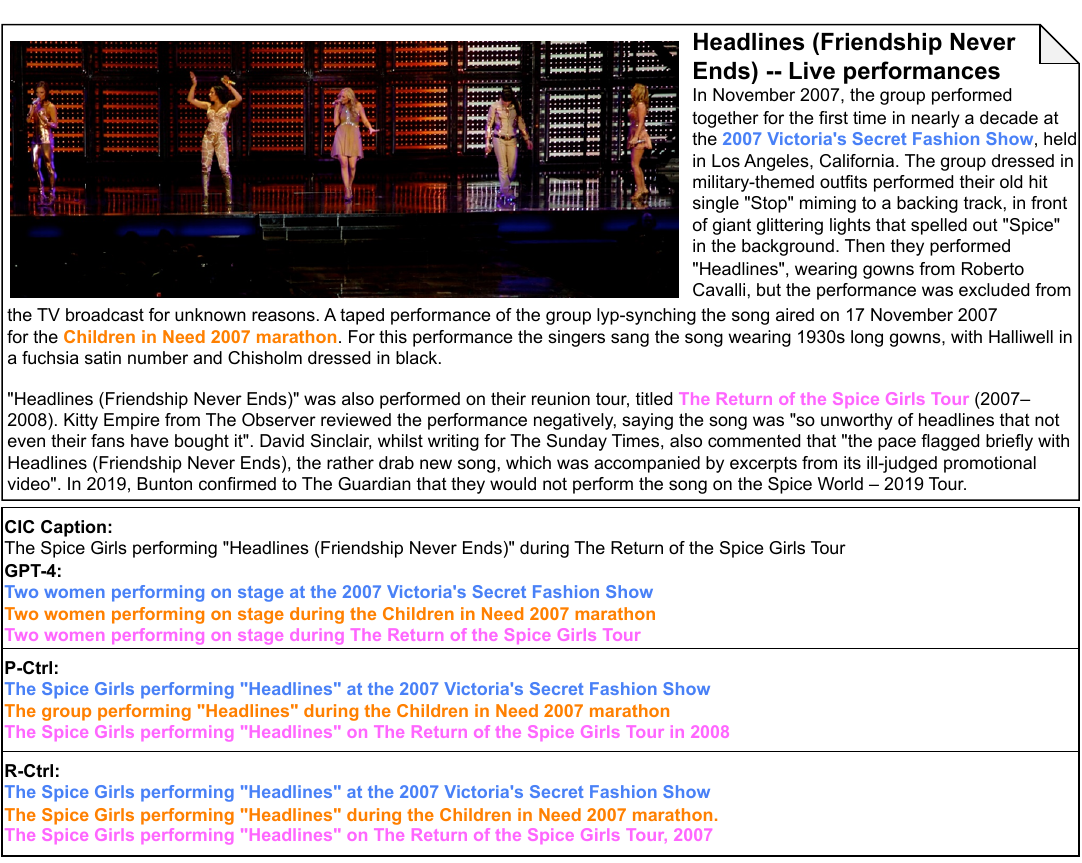}
    \caption{\CCICshort{} Captions Sample B}
    \label{fig:s2}
\end{figure*}

\begin{figure*}[!htb]
    \centering
    \includegraphics[width=\linewidth]{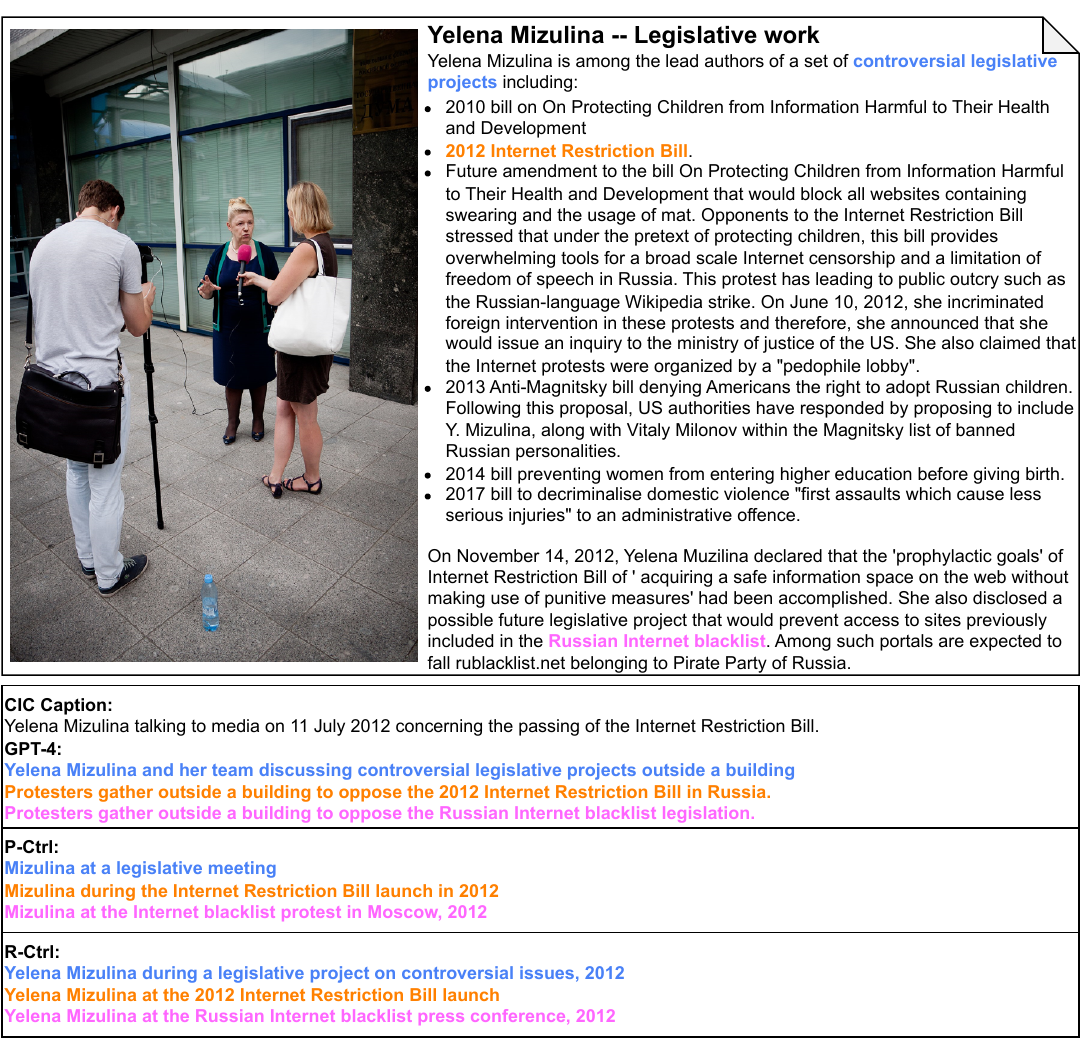}
    \caption{\CCICshort{} Captions Sample C}
    \label{fig:s3}
    \vspace{4cm}
\end{figure*}

\begin{figure*}[!htb]
    \centering
    \includegraphics[width=\linewidth]{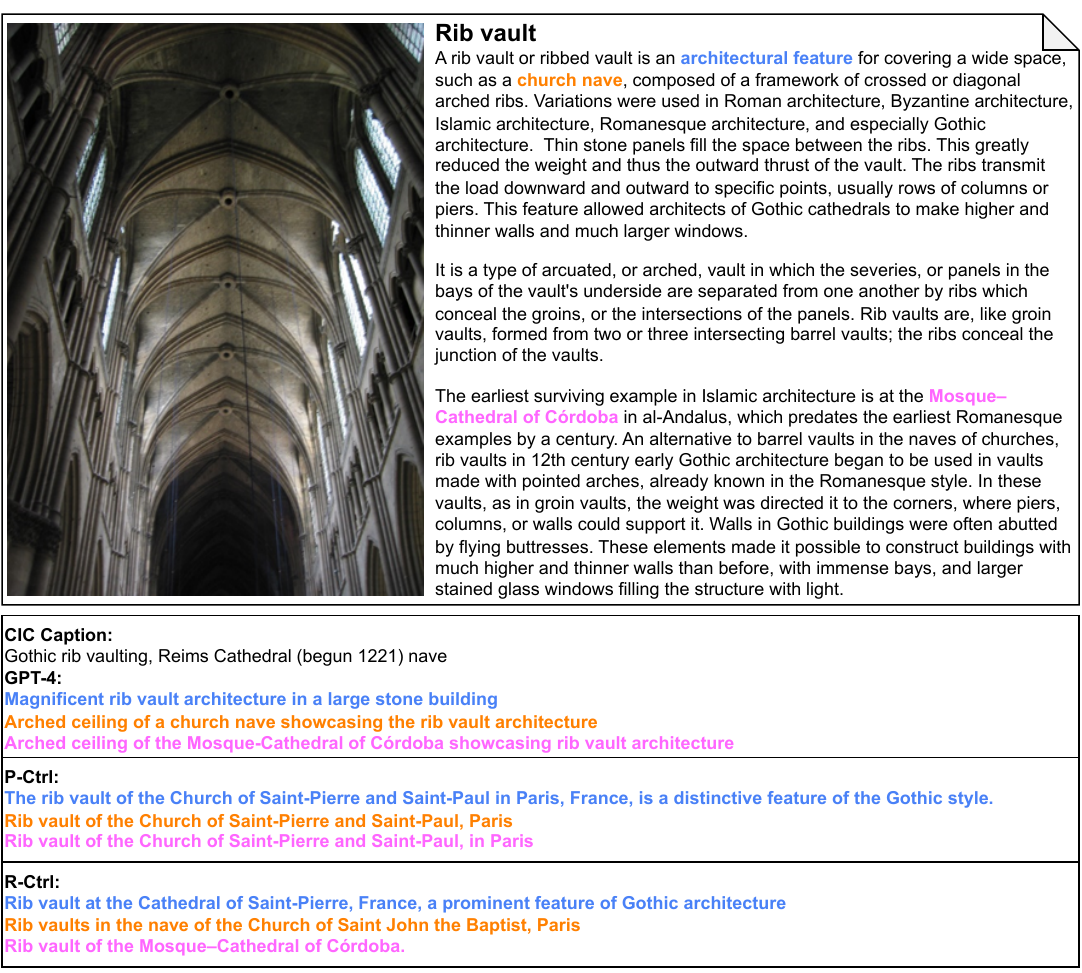}
    \caption{\CCICshort{} Captions Sample D}
    \label{fig:s4}
\end{figure*}

In this section, we demonstrate qualitative samples on the \CCICshort{} task to showcase the capabilities of proposed models at \CCICshort{}. The samples are shown in \cref{fig:s1,fig:s2,fig:s3,fig:s4}, where highlights and their respective \CCICshort{} captions are aligned in colors. It can be observed that each model endeavors to infer the image caption 
with a particular focus on the elements highlighted within the image.

\section{GPT-4 Prompts for Highlight Selection} 
\Cref{tab:prompt_highlight} demonstrates an example prompt and response from GPT-4 for selecting the highlight from context to construct evaluation samples.

\section{GPT-4 Prompts for \CCICshort{}} 
\Cref{tab:ccic_prompt} demonstrates an example prompt and response from GPT-4 for performing the \CCICshort{} task.

\section{GPT-4V Prompts for \CCICshort{} Evaluation}
\Cref{tab:ccic_eval_1,tab:ccic_eval_2,tab:ccic_eval_3} demonstrate an example prompt and response from GPT-4V for comparatively evaluating two \CCICshort{} captions. Note that images will also be given to the GPT-4V-empowered evaluator.

\section{GPT-4 Prompts for CIC} \label{sec::prompt}
\Cref{tab:prompt_cic} demonstrates an example prompt and response from GPT-4 for performing the CIC task.

\begin{table*}[ht]\centering
\caption{GPT-4 prompt for selecting highlights for \CCICshort{} evaluation}
    \label{tab:prompt_highlight}
\begin{minipage}{0.99\linewidth}\vspace{0mm}    \centering
\begin{tcolorbox}
    \centering
    \small
     \hspace{-6mm}
    \begin{tabular}{p{0.99\linewidth}}

\tbf{Prompt for Evaluation Highlights Selection}
\\
Task:\\
From the provided context, extract and highlight ten different informative keyphrases (or keywords) that contain valuable and relevant information. The extracted content can be a single word or phrases made up with several words. They should encapsulate key details, descriptive elements, or contextual insights that would be valuable for image captioning tasks. The aim is to ensure that some meaningful caption texts can be generated relevant to these context. Output these extracted keyphrases in one line, splitted by ``$|$" (i.e., in a restricted format as ``keyphrase1 $|$ keyphrase2 $|$ ..."; this requirement is very important and must be followed with no excuse) DIRECTLY (without any explanation or details).
\\\\
Context Section:
\\\\
Vermillion Township, Ashland County, Ohio. Vermillion Township, Ashland County, Ohio. Vermillion Township is one of the fifteen townships of Ashland County, Ohio, United States. The 2010 census found 2,618 people in the township, 2,170 of whom lived in the unincorporated portions of the township.
\\\\
Keyphrase Extraction:
\\
\midrule

\tbf{Response:} 
\\
Vermillion Township $|$ Ashland County $|$ Ohio $|$ fifteen townships $|$ 2010 census $|$ 2,618 people $|$ unincorporated portions $|$ United States $|$ 2,170 residents $|$ township population
\

    \end{tabular}
\end{tcolorbox}
    
\vspace{-2mm}
\end{minipage}
\end{table*}

\clearpage


\begin{table*}[ht]\centering
\caption{GPT-4 prompt for the \CCICshort{} task}
    \label{tab:ccic_prompt}
\begin{minipage}{0.99\linewidth}\vspace{0mm}    \centering
\begin{tcolorbox} 
    \centering
     \scriptsize
     \hspace{-6mm}
    \begin{tabular}{p{0.99\linewidth}}

\tbf{Prompt for the \CCICshort{} Task}
\\
Controllable Contextual Image Captioning Task\\
Objective: Your goal is to generate a short image caption that aligns with the provided context, particularly focusing on the highlighted segments.
\\\\
Instructions: \\\\
Context Section: Read the context provided to understand the overall theme or narrative. The context section will be given as the combination of a page title, a section title, and a section body. Some extra images other than the target image in the context section, if any, are included in the context section and described by [ImageCaptions] followed by text descriptions of the image to enrich the context.
\\\\
Image Description: Analyze the brief description or attributes of the image.
\\\\
Context Highlight: Pay special attention to the highlighted segments in the context. These are crucial cues for your caption. Highlights from different positions will be separated by line break.
\\\\
Caption Generation: Construct a one-sentence caption for the image that is both descriptive of the image and relevant to the context, especially the highlighted context cues. Output the caption DIRECTLY (one sentence without any explanation or details).
\\\\
Task Data:
\\\\
Context Section:
\\ 
\lbrack PageTitle\rbrack\space USS Vermont (BB-20) \lbrack SectionTitle\rbrack\space USS Vermont (BB-20) \lbrack SectionText\rbrack \space
USS Vermont (BB-20), a Connecticut-class battleship, was the second ship of the United States Navy named after the 14th state. She was the third member of the class, which included five other ships. The Connecticut-class ships were armed with a main battery of four 12-inch (305 mm) guns and had a top speed of 19 knots (35 km/h; 22 mph). Vermont was laid down in May 1904 at the Fore River shipyard and launched in August 1905. The ship entered service with the Atlantic Fleet in March 1907.

Shortly after she entered service, Vermont joined the Great White Fleet for its circumnavigation of the globe in 1908–1909. She took part in the international Hudson–Fulton Celebration in New York in 1909 and made trips to Europe in 1910 and 1913. Thereafter, the ship became involved in interventions in several Central American countries, including the United States occupation of Veracruz during the Mexican Revolution, where two of her crew earned the Medal of Honor. During the United States' participation in World War I from April 1917 to November 1918, Vermont served as a training ship for engine room personnel. From November 1918 to June 1919, she made a series of trips to return American soldiers from Europe before being decommissioned in June 1920. She was sold for scrap in November 1923 according to the terms of the Washington Naval Treaty.
\\\\
Image Description:\\
a ship with smoke coming out of it
\\\\
Context Highlight:\\
Connecticut-class battleship
\\\\
Controllable contextual caption (do not add quotations surrounding the response):

\\
\midrule

\tbf{Response:} 
\\
Connecticut-class battleship USS Vermont (BB-20) sailing with smoke billowing from its stacks.
\

    \end{tabular}
\end{tcolorbox}
    
\vspace{-2mm}

\end{minipage}
\end{table*}

\begin{table*}[ht]\centering
\caption{GPT-4V prompt for the \CCICshort{} evaluation}
    \label{tab:ccic_eval_1}
\vspace{-2mm}
    
\begin{minipage}{0.99\linewidth}\vspace{0mm}    \centering
\begin{tcolorbox} 
    \centering
    
     \scriptsize
     \hspace{-6mm}
    \begin{tabular}{p{0.99\linewidth}}

\tbf{Prompt for \CCICshort{} Evaluation}
\\
\begin{minipage}[c]{.99\textwidth}
  \centering
\includegraphics[width=0.36\linewidth]{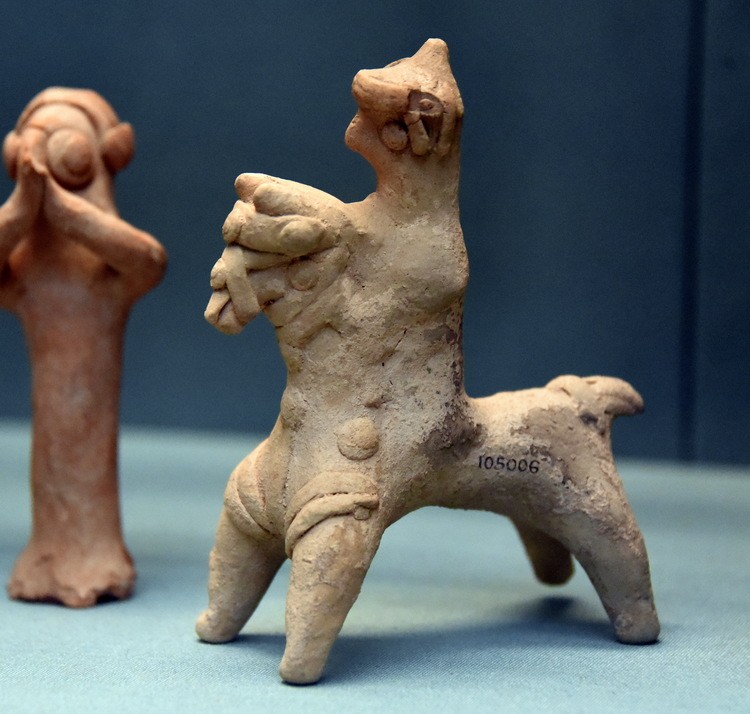}
\end{minipage} \\
We would like to request your feedback on the performance of two AI assistants in response to the controllable contextual image captioning task, where captions for an image will be generated based on the overall context and specific contextual highlights provided.
\\\\
Evaluation Steps:\\
1. You will be given the image, [Context Section] and [Highlighted Segments], followed by controllable contextual captions provided by the two assistant [ASSISTANT1 Caption] and [ASSISTANT2 Caption]: The context section will be given as the combination of a page title, a section title, and a section body. The highlighted segments are parts of the section body which will be given as words, phrases, or sentences, separated by line breaks. \\
2. You will thoroughly read the [Context Section] and [Highlighted Segments] provided, and carefully examine the [Image]. \\
3. You will read the caption generated by the AI assistant. \\
4. You will evaluate the controllable contextual image captioning quality of the two AI assistants, in terms of 4 aspects (which are "Relevance with Context", "Relevance with Highlight", "Consistency with Image", and "Overall Quality") - see below for individual criteria of these aspects. Each criterion should be considered in isolation to provide a clear and focused evaluation.\\
5. You will complete the following five sections IN ORDER (namely, [ASSISTANT1-Reasoning], [ASSISTANT2-Reasoning], [Comparison-Reasoning], [ASSISTANT1-Score], and [ASSISTANT2-Score])\\
\\\\
Evaluation Criteria:\\
- Relevance with Context: This metric rates how relevant the caption is to the given context. It assesses whether the caption pertains to and is appropriate for the contextual information provided, without necessarily reflecting the entire context. Captions should be scored based on their pertinence and the degree to which they relate to the context. Annotators should deduct points for captions that do not relate to or ignore the context. Higher scores should be awarded to captions that show a clear and significant relevance to the context.\\
- Relevance with Highlight: This metric evaluates how well the caption aligns with the highlighted segments provided. The caption should accurately reflect the information contained in the highlighted segments, ensuring that it is relevant and integrated into the overall caption. Annotators are advised to penalize captions that fail to address the highlighted segments or do so in a manner that does not give them adequate prominence or relevance.\\
- Consistency with Image: This metric evaluates the accuracy with which the caption represents elements or themes that are verifiably present in the image, based on the provided image descriptions. The caption should not introduce content that is clearly absent from the image. It needs to maintain a clear and direct connection to the key elements depicted in the image. Annotators should deduct points for captions that include inconsistencies or introduce elements not discernible in the image. Higher scores should be reserved for captions that are faithful to the image's visible content.\\
- Overall Quality: This metric assesses the caption's overall effectiveness in the CCIC task, emphasizing its coherence with the overall context, alignment with the image, and relevance to the highlighted segment. A high-quality caption should seamlessly integrate these elements, providing an accurate, informative, and engaging description of the image that resonates with the given context and highlights.\\
\tbf{To Be Continued Next Page}
    \end{tabular}
\end{tcolorbox}
    
\end{minipage}
\end{table*}

\begin{table*}[ht]\centering
\caption{GPT-4V prompt for \CCICshort{} evaluation}
    \label{tab:ccic_eval_2}
\begin{minipage}{0.99\linewidth}\vspace{0mm}    \centering
\begin{tcolorbox} 
    \centering
     \scriptsize
     \hspace{-6mm}
    \begin{tabular}{p{0.99\linewidth}}
HINT: \\
1. [ASSISTANT1-Reasoning] and [ASSISTANT2-Reasoning] will be used to record your reasoning and comments on the controllable contextual image caption generation quality of the two AI assistants, respectively; \\
2. [Comparison-Reasoning] will be used to record your feedback (with supporting evidence) for comparisions between the two AI assistants, which will be used to support the below two marking sections;\\
3. [ASSISTANT1-Score] and [ASSISTANT2-Score] will be used to record your controllable contextual image caption scores of the two AI assistants, respectively. Each assistant receives an integer score on a scale of 1 to 5 for each criteria, where a higher score indicates better performance according to the evaluation criteria.\\
4. Below is an example for requested output format of measuring one of the assistants:\\ 
\\
\text{[ASSISTANT1-Reasoning]} (*example):\\
- Relevance with Context: some feedback with supporting evidance...\\
- Relevance with Highlight: some feedback with supporting evidance...\\
- Consistency with Image: some feedback with supporting evidance...\\
- Overall Quality: some feedback with supporting evidance...\\
\\
\\
\\
\text{[ASSISTANT1-Score]} (*example):\\
- Relevance with Context: 3\\
- Relevance with Highlight: 1\\
- Consistency with Image: 4\\
- Overall: 2\\
\\
\\
\\
\text-----------------Evaluation Starts---------------------
\\
\\
\text{[Context Section]}:\\
\text{[PageTitle]} Euphrates Handmade Syrian Horses and Riders \text{[SectionTitle]} Euphrates Handmade Syrian Horses and Riders \text{[SectionText]} The Euphrates Handmade Syrian Horses and Riders (EU\_{HSHRs} ) are zoomorphic clay figurines representing horses and horses with riders. They date from the late Iron Age period (mid 8th–7th centuries BCE)and were produced in the Middle Euphrates region, alongside anthropomorphic figures known as Euphrates Syrian Pillar Figurines (EU\_SPFs).\\
\\
\text{[Highlighted Segments]}:\\
zoomorphic clay figurines\\
\\
\text{[ASSISTANT1 Caption]}:\\
EU\_HSHR  from Karkemish. The British Museum, London (museum no. 105006).\\
\\
\text{[ASSISTANT2 Caption]}:\\
Euphrates Handmade Syrian Horses and Riders, zoomorphic clay figurines\\
\\
\\
\\
Please make sure you read and understand these instructions carefully, and complete the following five sections IN ORDER: (1) firstly reasons via "[ASSISTANT1-Reasoning]:" and "[ASSISTANT2-Reasoning]:"; (2) secondly compares two assistants via "[Comparison-Reasoning]:"; and (3) finally marks them via "[ASSISTANT1-Score]:" and "[ASSISTANT2-Score]:".\\
\hrule
\vspace{1mm}
\tbf{To Be Continued Next Page}

    \end{tabular}
\end{tcolorbox}
    
\vspace{-2mm}
\end{minipage}
\end{table*}

\begin{table*}[ht]\centering
\caption{GPT-4V response in the \CCICshort{} evaluation}
    \label{tab:ccic_eval_3}
\begin{minipage}{0.99\linewidth}\vspace{0mm}    \centering
\begin{tcolorbox} 
    \centering
     \scriptsize
     \hspace{-6mm}
    \begin{tabular}{p{0.99\linewidth}}
\tbf{Response:}\\
\text{[ASSISTANT1-Reasoning]}:\\
- Relevance with Context: The caption provided by Assistant 1 includes the acronym EU\_HSHR, which is relevant to the context of Euphrates Handmade Syrian Horses and Riders. However, it does not mention the specific characteristic of being "zoomorphic clay figurines," which is a key aspect of the context.\\
- Relevance with Highlight: The caption fails to mention the highlighted segment "zoomorphic clay figurines," which is a significant oversight as this is a defining feature of the EU\_HSHRs.\\
- Consistency with Image: The caption accurately identifies the object in the image as an EU\_HSHR and provides additional information about its location (The British Museum, London) and museum number, which is consistent with the type of object depicted in the image.\\
- Overall Quality: The caption is informative and provides specific details about the object's museum location and identification number, but it lacks the integration of the highlighted segment, which affects the overall quality of the caption in the context of the task.\\
\\

\text{[ASSISTANT2-Reasoning]}:\\
- Relevance with Context: Assistant 2's caption directly references the context by repeating the title "Euphrates Handmade Syrian Horses and Riders" and includes the term "zoomorphic clay figurines," which is central to the context.\\
- Relevance with Highlight: The caption successfully incorporates the highlighted segment "zoomorphic clay figurines," directly addressing the key feature of the EU\_HSHRs.\\
- Consistency with Image: The caption is consistent with the image as it describes the object as a zoomorphic clay figurine, which is what is depicted.\\
- Overall Quality: The caption is coherent and relevant, effectively integrating the context and the highlighted segment with the image, resulting in a high-quality caption for the task.\\
\\
\text{[Comparison-Reasoning]}: \\
When comparing the two captions, Assistant 2's caption is superior in all aspects. It is more relevant to the context as it includes the full title and the term "zoomorphic clay figurines," which is the highlighted segment. Assistant 1's caption, while providing specific museum information, fails to mention the highlighted segment, which is a critical part of the task. Assistant 2's caption is also consistent with the image, as it describes exactly what is shown. Overall, Assistant 2's caption is more coherent and integrates all required elements of the task more effectively than Assistant 1's caption. \\
\\
\text{[ASSISTANT1-Score]}:\\
- Relevance with Context: 3\\
- Relevance with Highlight: 1\\
- Consistency with Image: 4\\
- Overall Quality: 2\\
\\
\text{[ASSISTANT2-Score]}: \\
- Relevance with Context: 5\\
- Relevance with Highlight: 5\\
- Consistency with Image: 5\\
- Overall Quality: 5\\
  \end{tabular}
\end{tcolorbox}
    
\vspace{-2mm}
\end{minipage}
\end{table*}

\begin{table*}[!ht]\centering
\caption{GPT-4 Prompt for the standard CIC Task}
    \label{tab:prompt_cic}
\begin{tcolorbox} 
    \centering
    \small
     \hspace{-6mm}
    \begin{tabular}{p{0.99\linewidth}}

\tbf{Prompt for CIC Task}
\\
CONTEXTUAL IMAGE CAPTIONING REQUEST
\\\\
TASK: Based on the context and image description provided, please generate a contextual image caption that captures the essence of the image and its significance in the given context. For your reference, the Context Section will be given as the combination of a page title, a section title, and a section body. Some extra images other than the target image in the context section, if any, are included in the context section and described by [ImageCaptions] followed by text descriptions of the image to enrich the context. The image to be captioned itself is not directly accessible by the model; instead, a textual description of the image is given in the ``Image Description" after the context section. Generate the proper caption for the image when the image is given together with the context. Please generate the caption directly without any explainations.
\\\\
Context Section:
[PageTitle] Hagerman horse [SectionTitle] Discovery [SectionText] A cattle rancher named Elmer Cook discovered some fossil bones on this land in Hagerman, Idaho. In 1928, he showed them to Dr. H. T. Stearns of the U.S. Geological Survey who then passed them on to Dr. James W. Gidley at the Smithsonian Institution. Identified as bones belonging to an extinct horse, the area where the fossils were discovered, called the Hagerman Horse Quarry, was excavated and three tons of specimens were sent back to the Smithsonian in Washington, D.C.
Excavation of the fossils continued into the early 1930s. The Hagerman Horse Quarry floor grew to 5,000 square feet (460 $m^2$) with a backwall 45 feet (14 m) high. Ultimately five nearly complete skeletons, more than 100 skulls, and forty-eight lower jaws as well as numerous isolated bones were found. Some paleontologists believed that such a large amount of fossils found in one location was because of the quarry area being a watering hole at one point. The waterhole could have been where the bones of the Hagerman horses accumulated as injured, old, and ill animals, drawn to water, died there. Other paleontologists think that an entire herd of these animals drowned attempting to ford a flooded river and were swept away in the current and ended up buried in the soft sand at the bottom.
\\\\
Image Description:
a zebra with a white background
\\\\
Contextual Caption (do not add quotations surrounding the response):
\\
\midrule

\tbf{Response:} 
\\
Zebra-like ancestor of the Hagerman horse, symbolizing its evolutionary connection
\

    \end{tabular}
\end{tcolorbox}
    
\vspace{-2mm}
\end{table*}
\clearpage

\end{document}